\documentclass[preprint,12pt]{elsarticle}   

\usepackage[utf8]{inputenc}
\usepackage{geometry}
\usepackage{graphicx}
\usepackage{xcolor}
\usepackage{booktabs}
\usepackage{multirow}
\usepackage{tabularx}
\usepackage{array}
\usepackage{longtable}
\usepackage{ragged2e}
\usepackage{amsmath}
\usepackage{amssymb}
\usepackage{mathtools}
\usepackage{bm}
\usepackage{pifont}
\usepackage{caption}
\usepackage[section]{placeins}
\usepackage[switch]{lineno}
\usepackage{fancyhdr}
\usepackage{hyperref}
\hypersetup{
    colorlinks=true,
    linkcolor=blue,
    citecolor=green,
    urlcolor=blue
}
\fancypagestyle{plain}{%
    \fancyhf{}
    \fancyhead[L]{\textit{Owais et al.}}
    \fancyhead[R]{\textit{Recent Advances in AI-Driven Plant Phenotyping}}
    \fancyfoot[C]{\thepage}
    
}

\newcommand{\cmark}{\textcolor{green}{\ding{51}}}
\newcommand{\xmark}{\textcolor{red}{\ding{55}}}
\newcommand{\pmark}{\textcolor{orange}{\ding{108}}}
\newcolumntype{P}[1]{>{\RaggedRight\arraybackslash}p{#1}}
\newcolumntype{Q}{>{\RaggedRight\arraybackslash}X}
\newcolumntype{M}[1]{>{\centering\arraybackslash}p{#1}}
\newcolumntype{J}[1]{>{\justifying\arraybackslash}p{#1}}

\begin{document}

\begin{frontmatter}

\title{Recent Advances in Agentic Agri-Robotic Phenotyping: A Perspective Review from Fragmented Multimodal Sensing to Unified PhenoAgent Intelligence}

\author[a]{Muhammad Owais\corref{cor1}} 
\author[b]{Ehtesham Iqbal} 
\author[b]{Samee Ullah Khan} 
\author[d]{Muhammad Umraiz} 
\author[b,c]{Yusra Abdulrahman} 
\author[a]{Irfan Hussain\corref{cor1}} 

\address[a]{Khalifa University Center for Autonomous Robotic Systems (KU-CARS), Khalifa University of Science and Technology, Abu Dhabi, United Arab Emirates.}
\address[b]{Advanced Research and Innovation Center (ARIC), Khalifa University of Science and Technology, Abu Dhabi, United Arab Emirates.}
\address[c]{Department of Aerospace Engineering, Khalifa University of Science and Technology, Abu Dhabi, United Arab Emirates.}
\address[d]{Department of Computer Science and Engineering, Chung-Ang University, Seoul, South Korea.}

\cortext[cor1]{Corresponding Authors: muhammad.owais@ku.ac.ae and irfan.hussain@ku.ac.ae}

\begin{abstract}
\small
\itshape

This review examines the evolution of plant phenotyping from conventional manual trait measurement to high-throughput, robotic, and artificial intelligence-driven crop monitoring. Despite significant advances in imaging, autonomous platforms, multimodal sensing, and deep learning, current phenotyping systems remain fragmented across sensing modalities, crop traits, growth stages, environments, and management objectives. We therefore frame phenotyping as an integrated \emph{seed-soil-plant-environment-management} (SSPEM) intelligence problem, where crop performance reflects interactions among seed quality, root-zone conditions, plant development, environmental exposure, and management actions. The review synthesizes conventional, high-throughput, robotic, and AI-driven phenotyping approaches, highlighting their capabilities and persistent limitations in temporal integration, multimodal reasoning, biological interpretation, and actionable decision support. Building on this analysis, we introduce a conceptual PhenoAgent framework that extends phenotyping beyond the estimation of isolated traits to evidence-based crop-state interpretation, uncertainty-aware reasoning, and management-oriented support. The PhenoAgent concept primarily brings together scattered advances in phenotyping to deliver insights ranging from detailed to high-level, such as what is happening in the crop, why it might be occurring, what evidence is missing, and what actions or additional measurements should be considered. We also discuss challenges in dataset scarcity, annotation, benchmarking, model generalization, and explainability. By linking multimodal phenotyping with agentic AI and closed-loop decision support, this review outlines a path to interpretable, scalable, and deployment-oriented crop intelligence. Additional resources related to this review, including supplementary materials, curated references, datasets, tools, and future updates, will be made available through the \href{https://github.com/Owais-CodeHub/PhenoAgent}{GitHub} repository.

\end{abstract}

\end{frontmatter}

\clearpage
\begingroup
\hypersetup{linkcolor=blue}
\small
\tableofcontents
\endgroup
\clearpage

\section{Introduction}
Global agriculture is entering a period in which crop productivity must increase while water, land, labor, fertilizer, and climate stability become more constrained. Long term food demand analyzes indicate that global crop production must increase substantially to satisfy population growth and changing diets, yet yield trends in several major crops remain insufficient without faster innovation in breeding, agronomy and production systems \cite{tilman2011global,ray2013yield}. These challenges are particularly acute in arid and semi arid regions, where heat, salinity, water scarcity, protected cultivation, and dependence on imported food create strong pressure to develop crops and management strategies that perform reliably under resource limited conditions. In such environments, improvement cannot depend only on genetic selection or climate control hardware. It also requires the ability to continuously measure crop response, understand why performance differs between seed lots, substrates, environments, and management regimes, and convert this evidence into timely decisions.
Plant phenotyping provides the measurable bridge between genotype, environment, management, and crop outcome. It quantifies visible and hidden traits such as germination, emergence, root zone response, morphology, physiology, stress, disease, growth dynamics, quality, and yield. Although genomics, sensors, robotics, and AI have advanced rapidly, the ability to measure biologically meaningful phenotypes at sufficient scale, frequency, and decision relevance remains a major bottleneck in crop improvement and controlled environment agriculture \cite{furbank2011phenomics,fiorani2013future}. This bottleneck is no longer only a question of how many plants can be measured. The deeper challenge is how to understand phenotype formation across the full seed soil plant environment management (SSPEM) continuum, where seed vigor, substrate salinity, moisture distribution, fertigation timing, light exposure, genotype sensitivity, and disease pressure may interact to determine final productivity.

\begin{figure}[h!]
    \centering
    \includegraphics[width=1\textwidth]{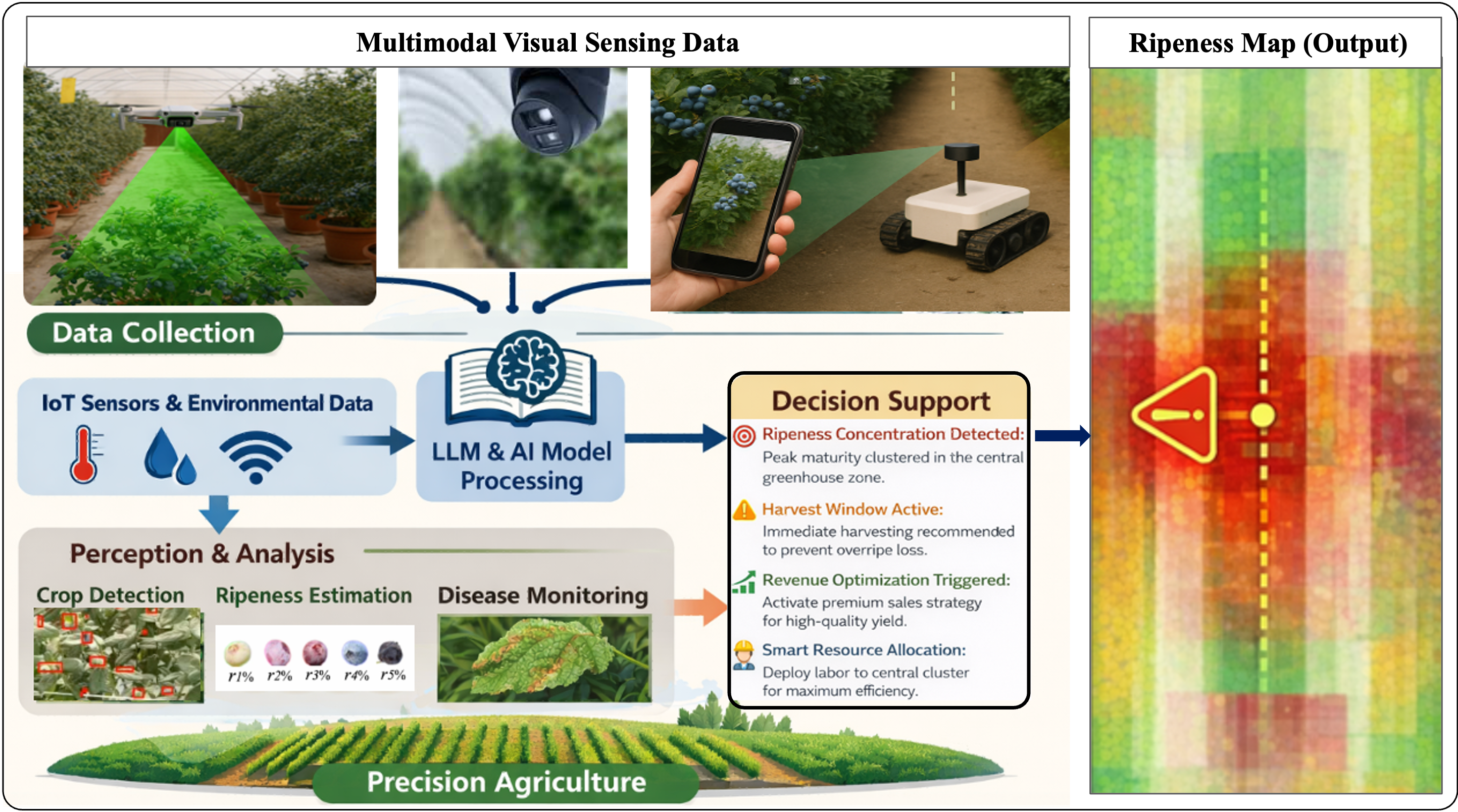}
\caption{Conceptual framework for AI-driven precision agriculture integrating multimodal visual sensing with IoT and environmental data. AI models and large language models (LLMs) support crop detection, ripeness estimation, and disease monitoring, producing spatial ripeness maps and decision support for harvest scheduling, resource allocation, and revenue optimization.}
\label{fig:phenoagent_ripeness_mapping}
\end{figure}

\subsection{Plant Phenotyping as a Bottleneck in Crop Improvement}

In breeding, phenotyping determines whether genetic variation is expressed as useful field or greenhouse performance. In agronomy, it supports decisions on seed selection, irrigation, fertigation, substrate design, stress management, disease control, and harvest timing. In controlled environment agriculture, phenotyping can help optimize lighting, climate, CO2, humidity, nutrient delivery, root zone condition, and crop scheduling. Across these contexts, the phenotype is dynamic and conditional. The same genotype may behave differently across environments, and the same environment may produce different outcomes depending on management. This explains why high throughput phenotyping has repeatedly been described as a key frontier for crop improvement \cite{araus2014field,tardieu2017plant}.
The bottleneck is especially severe because many important crop responses unfold over time and across biological layers. A seed lot may pass a standard germination test but later produce non uniform seedlings under salinity or heat. A plant may appear healthy in RGB imagery while root zone electrical conductivity or moisture distribution is already driving stress. A greenhouse dashboard may show acceptable average temperature and humidity while individual trays experience local stress due to airflow, irrigation distribution, substrate heterogeneity, or genotype sensitivity. These examples show why phenotyping must move beyond isolated trait snapshots toward longitudinal crop state interpretation. Data standards such as MIAPPE and phenotyping data integration frameworks provide important foundations, but many AI pipelines still disconnect images, sensor logs, management events, and biological metadata \cite{coppens2017unlocking,papoutsoglou2020miappe}.

Controlled environment agriculture creates an especially strong case for next generation phenotyping because it combines rich sensing with controllable interventions. Smart greenhouses can collect data from cameras, IoT devices, fertigation systems, climate controllers, robotic scouts, and operational records, yet these data often remain separated in different software layers. Digital twin and smart farming studies show that crop monitoring, simulation, and decision support are becoming more feasible, but they also reveal a gap between descriptive dashboards and biologically grounded recommendations \cite{wolfert2017big,purcell2023digital,rahman2024digitaltwin}. A next generation phenotyping system should therefore maintain a temporal memory of crop development and use it to explain deviations, support decisions, and learn from intervention outcomes. Figure~\ref{fig:phenoagent_ripeness_mapping} presents a representative example within the broader PhenoAgent ecosystem, where ripeness-based harvesting and disease assessment illustrate how visual crop observations can be transformed into spatial information on crop condition to support harvesting decisions and management planning.

\subsection{From Trait Measurement to Crop Intelligence}

Several recent reviews have made important contributions to plant phenotyping. Historical and phenomics reviews have clarified the phenotyping bottleneck and the need to transform sensor data into knowledge \cite{furbank2011phenomics,pieruschka2019plant}. Image based and deep learning reviews have summarized progress in detection, segmentation, classification, disease monitoring, trait extraction, and yield prediction \cite{jiang2020deep,murphy2024deep}. Seed focused reviews have highlighted the growing role of AI in high throughput seed phenotyping, while robotic and sensor focused reviews have mapped UAV, UGV, fixed camera, and multimodal sensing platforms \cite{jin2025seed,atefi2021robotic,zhang2026ground}. More recently, foundation model and digital twin reviews have expanded the discussion toward reusable representations, smart agriculture, and decision oriented systems \cite{li2024foundation,peladarinos2023digital}. Recent multimodal LLM benchmarks in agriculture further support the transition from visual perception to agronomic reasoning. AgEval evaluates zero shot and few shot multimodal LLM performance across plant stress phenotyping tasks, showing that model reliability varies substantially across stress classes and improves with carefully selected examples~\cite{arshad2025leveragingvisionlanguagemodels}. PlantXpert extends this direction by evaluating evidence grounded multimodal reasoning for soybean and cotton phenotyping, highlighting that quantitative reasoning, biological grounding, and cross crop generalization remain challenging even for recent vision language models~\cite{wu2026uavimageryagronomicreasoning}. These findings reinforce the need for PhenoAgent like systems that combine perception models with crop knowledge, uncertainty estimation, temporal context, and expert validation.
Despite this progress, the literature remains fragmented across technology categories. Many studies focus on how to detect an organ, estimate a trait, classify a disease, navigate a robot, fly a UAV, build a dashboard, or train a model. These advances are necessary, but they do not fully address the broader phenotyping question. For instance, how can a system understand crop state as the outcome of linked seed, soil or substrate, root zone, plant, environment, genotype, and management evidence? This review addresses that gap by treating phenotyping as a crop intelligence problem rather than only a measurement problem.
\cite{bommasani2021foundation,lewis2020rag,li2024foundation}. 

\begin{figure}[h!]
\centering
\includegraphics[width=1\textwidth]{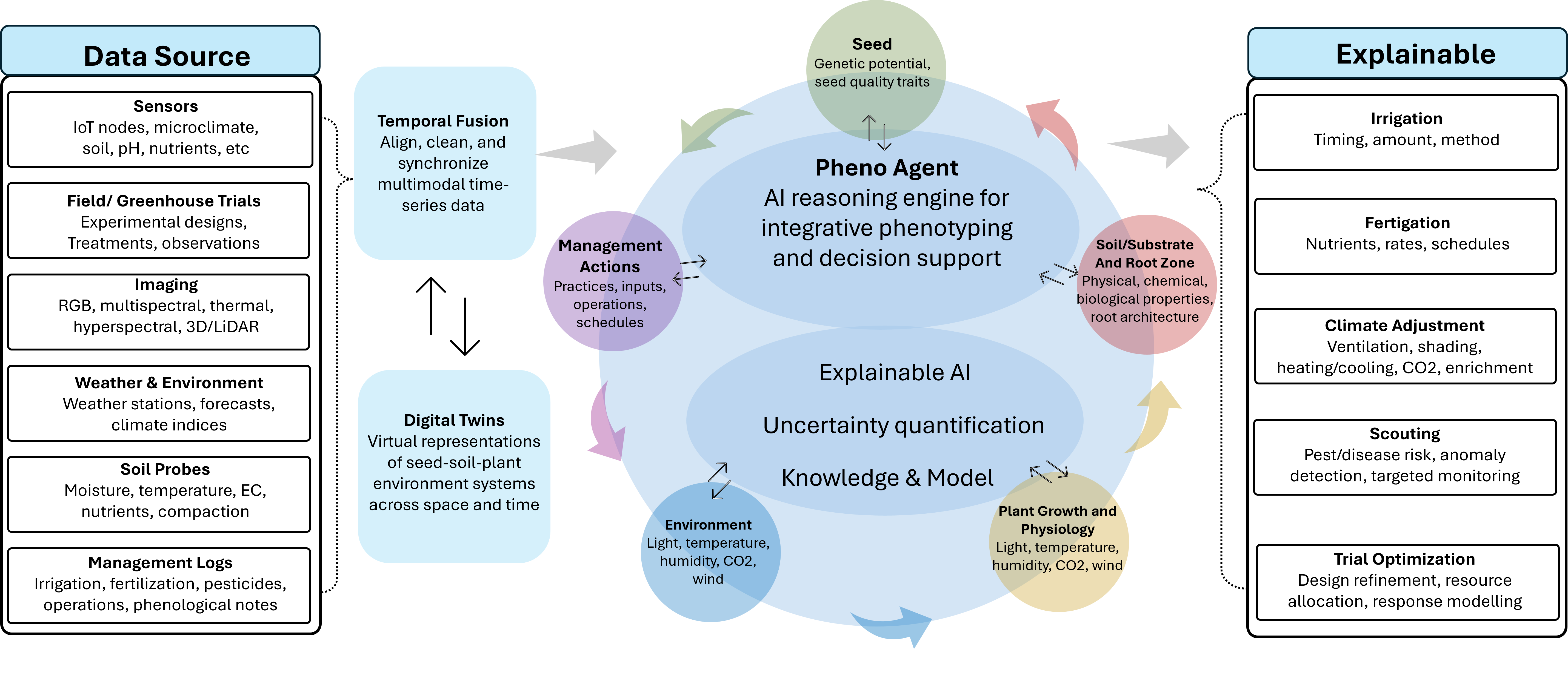}
\caption{Conceptual architecture of Pheno Agent for integrative phenotyping and decision support. Temporal fusion and digital twins support the integration of multimodal data across seed traits, soil and root-zone properties, plant growth and physiology, environmental conditions, and management practices. The framework combines AI reasoning, domain knowledge, and models with explainable AI and uncertainty quantification to support interpretable decisions.}
\label{fig:seed_soil_plant_continuum}
\end{figure}

This review is driven by the perspective that plant phenotyping is advancing through three partly overlapping phases. The first is conventional phenotyping, in which biological traits and reference measurements are determined via manual, destructive, or otherwise low-throughput methods, frequently incurring substantial costs in time, labor, scalability, and consistency. The second is high-throughput and AI-enabled phenotyping, where RGB, thermal, multispectral, hyperspectral, depth, LiDAR, IoT, UAV, UGV, fixed-camera, robotic, deep learning, foundation-model, multimodal-fusion, digital-twin, and dashboard technologies increase the speed and scale of crop monitoring \cite{tardieu2017plant,murphy2024deep,wang2025sensors}. The third phase is now taking shape as agentic phenotyping, where systems go beyond trait detection and sensor visualization to reason over temporal multimodal evidence, interpret probable stress causes, flag missing information, predict crop trajectories, and enable expert-validated interventions. In this review, we use \textit{PhenoAgent} to refer to this next-generation direction. Figure~\ref{fig:seed_soil_plant_continuum} depicts the SSPEM continuum that underpins PhenoAgent-based phenotyping.

\begin{table}[t]
\centering
\caption{Comparison of closely related AI-driven plant phenotyping review papers with our conceptual PhenoAgent review. 
\cmark~= addressed in detail; \pmark~= partially addressed; \xmark~= not a central focus.
PP~= plant phenotyping; SP~= seed phenotyping; RZ~= root zone; Rob.~= robotics; DL~= deep learning; FM~= foundation models; DT~= digital twins; AR~= agentic reasoning; CLA~= closed-loop action.
The comparison highlights that existing reviews strongly cover sensors, imaging, robotics, and deep learning, while a unified agentic seed-soil-plant-environment-management (SSPEM) intelligence framework remains underdeveloped.}
\label{tab:intro_review_comparison}
\tiny
\setlength{\tabcolsep}{1.2pt}
\renewcommand{\arraystretch}{1.12}
\resizebox{\linewidth}{!}{%
\begin{tabular}{P{2.6cm} J{2.55cm} M{0.55cm} M{0.55cm} M{0.60cm} M{0.55cm} M{0.55cm} M{0.60cm} M{0.55cm} M{0.60cm} M{0.60cm} J{2.9cm}}
\toprule
\textbf{Review} 
& \textbf{Focus} 
& \textbf{PP} 
& \textbf{SP} 
& \textbf{RZ} 
& \textbf{Rob} 
& \textbf{DL} 
& \textbf{FM} 
& \textbf{DT} 
& \textbf{AR} 
& \textbf{CLA} 
& \textbf{Limitations} \\
\midrule
Pieruschka and Schurr~\cite{pieruschka2019plant} & \noindent History, infrastructure, standardization, and future needs in plant phenotyping & \cmark & \pmark & \pmark & \pmark & \pmark & \xmark & \xmark & \xmark & \xmark & \noindent Does not focus on agentic AI, foundation models, or closed loop seed soil plant reasoning. \\
Murphy et al.~\cite{murphy2024deep} & \noindent Deep learning in image based plant phenotyping & \cmark & \xmark & \xmark & \pmark & \cmark & \pmark & \xmark & \xmark & \xmark & \noindent Strong AI review, but primarily image based and not organized around multimodal decision intelligence. \\
Jin et al.~\cite{jin2025seed} & \noindent Deep learning for high throughput seed phenotyping & \pmark & \cmark & \xmark & \xmark & \cmark & \xmark & \xmark & \xmark & \xmark & \noindent Seed specific focus, with limited connection to downstream plant, environment, management, and intervention decisions. \\
Wang et al.~\cite{wang2025sensors} & \noindent Image based high throughput phenotyping technologies and trends & \cmark & \pmark & \pmark & \cmark & \cmark & \pmark & \pmark & \xmark & \pmark & \noindent Broad technology synthesis, but agentic reasoning and unified management intelligence are not central. \\
Zhang et al.~\cite{zhang2026ground} & \noindent Ground mobile robots for high throughput plant phenotyping from perception decision action perspective & \cmark & \xmark & \pmark & \cmark & \cmark & \pmark & \xmark & \pmark & \cmark & \noindent Strong robotic pipeline focus, but less emphasis on seed soil plant digital twins and foundation agent reasoning. \\
Li et al.~\cite{li2024foundation} & \noindent Foundation models in smart agriculture & \pmark & \xmark & \pmark & \pmark & \cmark & \cmark & \pmark & \pmark & \pmark & \noindent Broad smart agriculture focus, not specifically centered on plant phenotyping workflows and phenotypic validation. \\
\textbf{PhenoAgent (Our)} & \noindent Agentic multimodal seed soil plant environment management phenotyping intelligence & \cmark & \cmark & \cmark & \cmark & \cmark & \cmark & \cmark & \cmark & \cmark & \noindent Requires substantial computational resources \\
\bottomrule
\end{tabular}}
\end{table}
Table~\ref{tab:intro_review_comparison} clarifies the novelty of this review. The goal is not to provide another sensor catalogue or deep learning survey. Instead, the review synthesizes conventional phenotyping, current AI driven phenotyping, and next generation agentic AI into one conceptual pathway. It argues that phenotyping should evolve toward systems that can observe, integrate, explain, decide, and adapt. This perspective is particularly important for greenhouse and field trail workflows where the value of phenotyping depends not only on detecting traits, but on improving decisions about seed lots, substrates, irrigation, fertigation, climate control, scouting, and breeding selection.

\subsection{Review Scope and Literature Selection (PRISMA)}

\begin{figure}[h!]
    \centering
    \includegraphics[width=0.7\textwidth]{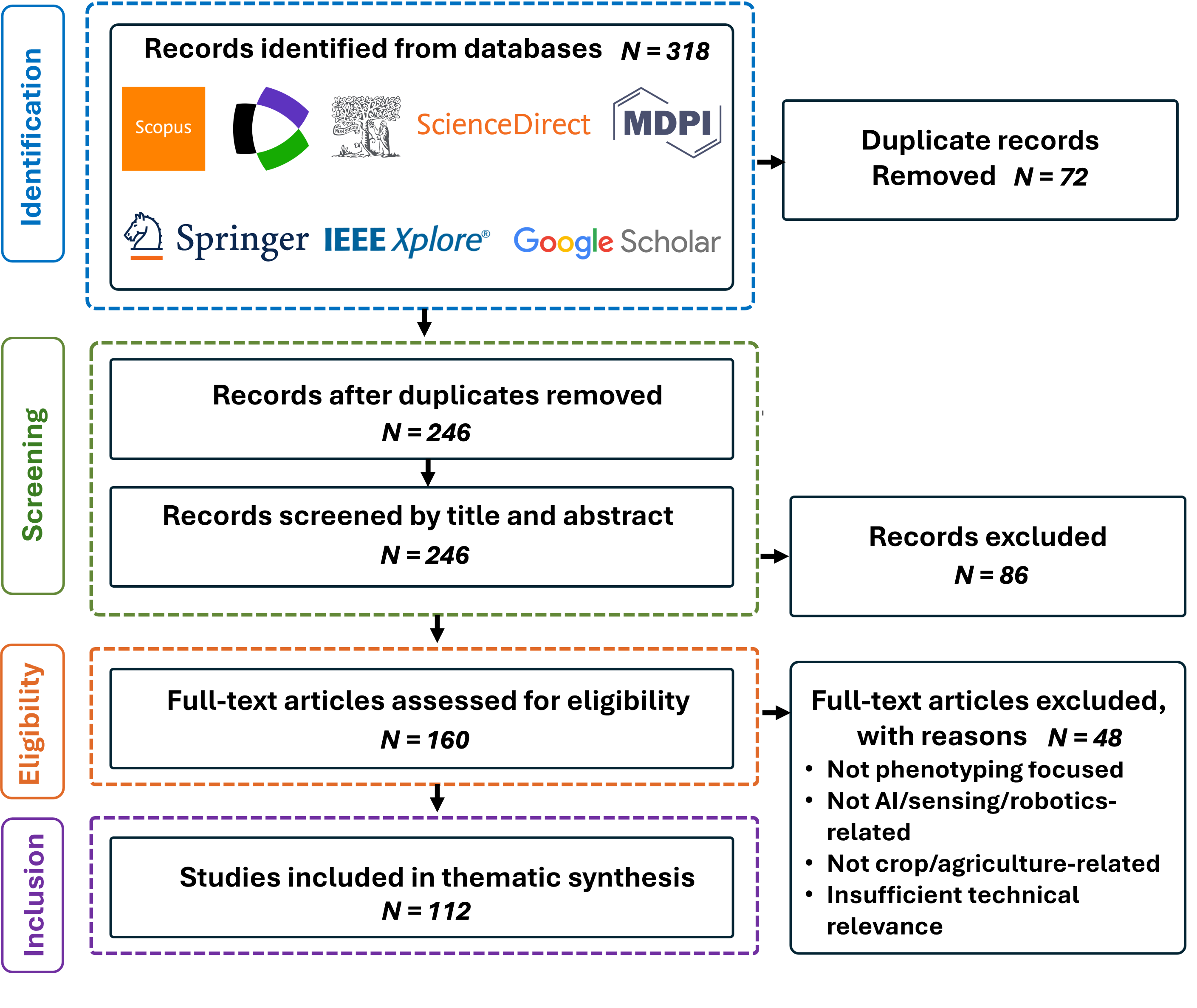}
    \caption{Literature selection PRISMA diagram illustrating the identification, screening, eligibility assessment, and inclusion stages. Of the 318 records initially identified, 246 underwent screening, 160 were assessed for eligibility, and 114 were included in the final review.}
    \label{fig:prisma_literature_selection}
\end{figure}

This review uses a perspective-oriented synthesis with a structured literature selection process. A formal meta-analysis was unsuitable because the topic spans multiple communities-plant phenotyping, seed and root-zone assessment, remote sensing, robotics, computer vision, deep learning, digital twins, smart greenhouses, foundation models, retrieval-augmented systems, and technology translation. The review therefore highlights technical progress and conceptual gaps across these connected areas. The literature search used a PRISMA-style workflow for transparent study identification and screening \cite{page2021prisma}. Records from major databases and publisher platforms were deduplicated, screened by title/abstract, assessed for full-text eligibility, and thematically grouped. Studies were included if they directly addressed plant phenotyping, AI-enabled sensing, robotic/automated data acquisition, multimodal crop analysis, digital crop intelligence, or agentic decision support; studies outside crop/agriculture, unrelated to phenotyping, or lacking technical relevance were excluded. Figure~\ref{fig:prisma_literature_selection} summarizes the workflow and the studies retained for synthesis.

The search strategy covers three linked domains. The first domain is conventional phenotyping, including manual, destructive, seed, germination, soil substrate, root zone, greenhouse, field trail, and genotype environment management phenotyping. The second domain is current AI enabled phenotyping, including RGB, thermal, multispectral, hyperspectral, depth, LiDAR, IoT, UAV, UGV, fixed camera, robotic, deep learning, transformer, promptable model, multimodal fusion, digital twin, and dashboard systems. The third domain is next generation crop intelligence, including foundation models, retrieval augmented generation, agentic AI, Human in the loop decision support, edge cloud deployment, high TRL translation, and commercialization. This structure mirrors the organization of the review and prevents the literature synthesis from becoming a loose collection of technologies.

\begin{table}[h!]
\centering
\caption{Research questions and corresponding contributions of the proposed review.}
\label{tab:intro_rq_contributions}
\scriptsize
\setlength{\tabcolsep}{3pt}
\renewcommand{\arraystretch}{1.16}
\begin{tabular}{P{1.2cm} P{6.1cm} J{8.5cm}}
\toprule
\textbf{RQ} & \textbf{Research question} & \textbf{Review contribution} \\
\midrule
RQ1 & How has plant phenotyping evolved from manual measurement to AI enabled automation? & \noindent Provides a compact historical synthesis of conventional phenotyping, high throughput sensing, robotic platforms, and AI driven trait extraction. \\
RQ2 & Which current technologies enable automated phenotyping? & \noindent Compares imaging, IoT, UAV, UGV, robotic, deep learning, foundation model, multimodal fusion, digital twin, and dashboard advances. \\
RQ3 & Where are current systems still fragmented? & \noindent Identifies gaps in dataset standardization, multimodal fusion, temporal reasoning, biological interpretability, and seed soil plant integration. \\
RQ4 & How can PhenoAgent define the next generation of phenotyping? & \noindent  Introduces a conceptual framework for agentic (SSPEM) intelligence with explanation, forecasting, and intervention recommendation. \\
RQ5 & What is needed for high TRL translation? & \noindent Discusses benchmarking, validation, edge/cloud deployment, human oversight, data ownership, commercialization, and operational adoption. \\
\bottomrule
\end{tabular}
\end{table}

\subsection{Research Questions and Contributions}

The aim of this review is to provide a focused and forward looking synthesis of automated plant phenotyping from conventional trait measurement to PhenoAgent based multimodal crop intelligence. The review is guided by the view that the next generation of phenotyping should not stop at image segmentation, vegetation indices, sensor dashboards, or yield prediction. Instead, it should connect observations across the (SSPEM) continuum and support explainable decisions that improve breeding, field trials, controlled environment agriculture, and commercial crop production. This aim is aligned with earlier calls to move from sensors to knowledge, but extends the discussion toward agentic AI, digital plant twins, human guided reasoning, closed loop action, and high TRL translation \cite{tardieu2017plant,lewis2020rag,rahman2024digitaltwin}. The review addresses five research questions as mentioned in Table \ref{tab:intro_rq_contributions}. 



\section{Conventional Phenotyping Across the Seed Soil Plant Continuum}
\label{sec-2}

\begin{figure}[h!]
    \centering
    \includegraphics[width=1\textwidth]{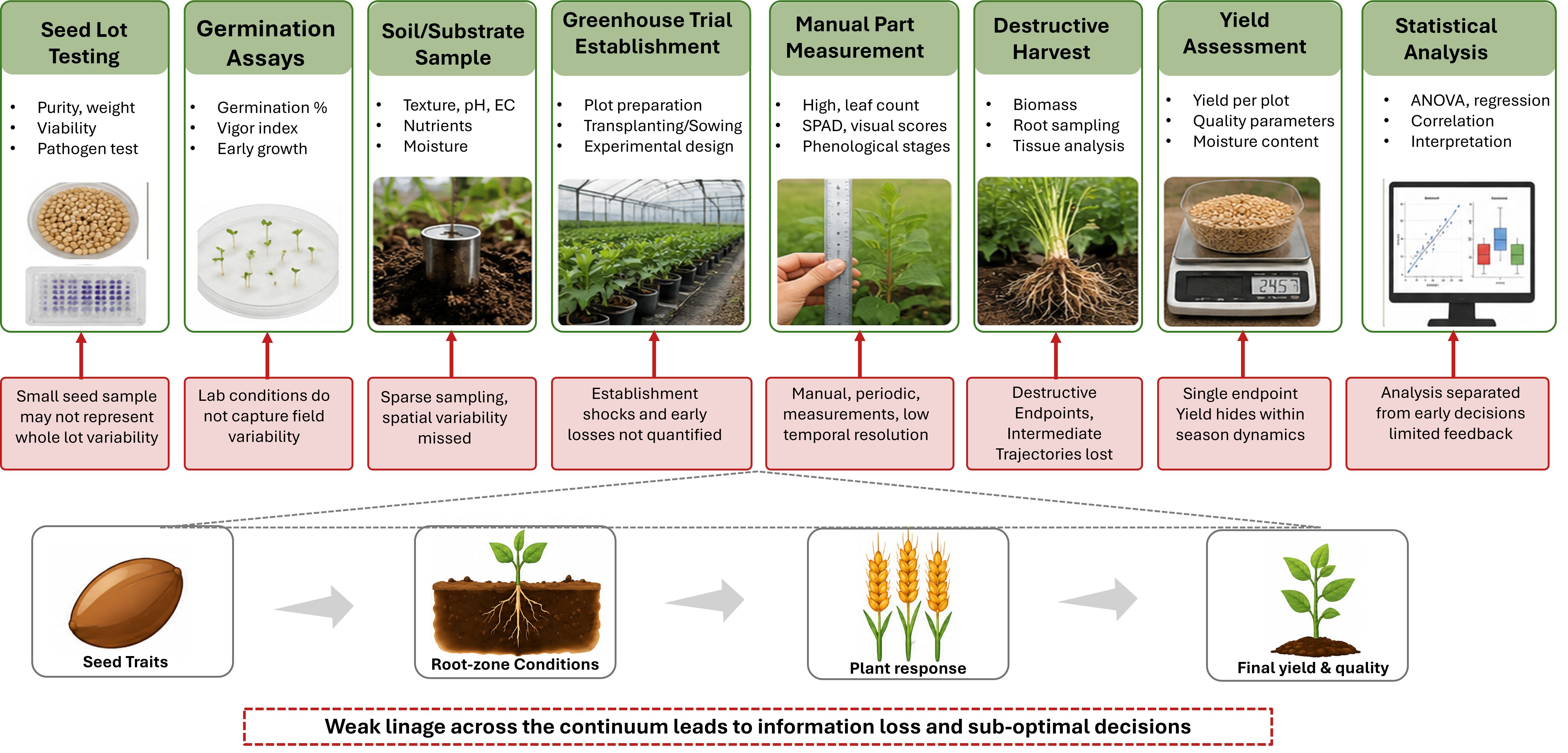}
    \caption{Conventional plant phenotyping workflow and its limitations. Sequential stages span seed lot testing, germination assays, soil/substrate sampling, greenhouse trial establishment, manual measurements, destructive harvest, yield assessment, and statistical analysis. Highlighted limitations include limited sample representativeness, unobserved spatial and temporal variability, unquantified establishment losses, and the loss of intermediate growth trajectories. Endpoint-focused assessments and delayed analysis constrain the integration of seed, soil, and plant responses and limit feedback for early decision-making.}
    \label{fig:conventional_seed_soil_plant_workflow}
\end{figure}

Conventional plant phenotyping refers to the long established set of visual, manual, destructive, laboratory, greenhouse, and field trial measurements used to evaluate seed quality, germination, soil or substrate conditions, plant growth, stress response, and yield. These methods remain scientifically important because they define the reference traits, biological terminology, and validation standards against which newer AI driven phenotyping systems must be evaluated. At the same time, conventional phenotyping is increasingly recognized as a bottleneck because crop improvement now requires faster, more frequent, more spatially resolved, and more biologically integrated measurements than manual protocols can usually provide \cite{furbank2011phenomics,fiorani2013future}. Figure \ref{fig:conventional_seed_soil_plant_workflow} summarizes conventional phenotyping workflow across the seed soil plant continuum. This section reviews conventional phenotyping across the seed soil plant continuum and highlights why its limitations motivate the transition toward multimodal, automated, and PhenoAgent based phenotyping.

\subsection{Manual and Destructive Phenotyping}

Manual phenotyping has historically been the backbone of crop science because it provides direct, interpretable, and biologically grounded measurements. Researchers and breeders have relied on traits such as plant height, leaf number, leaf area, stem diameter, tiller number, flowering time, biomass, root mass, harvest index, fruit number, grain yield, and visual stress scores to characterize genotypes and treatments. Many of these measurements are inexpensive and require limited instrumentation, but they are also labor intensive, observer dependent, and difficult to repeat at high temporal resolution. The phenotyping bottleneck emerged because genotyping and molecular tools accelerated faster than the ability to measure whole plant performance under relevant environmental and management conditions \cite{furbank2011phenomics,araus2014field}. Thus, the main constraint in many breeding and agronomic pipelines is no longer the availability of genetic variation, but the capacity to measure how that variation becomes a dynamic phenotype.

Destructive phenotyping remains especially important for traits that cannot be reliably inferred from external appearance alone. Biomass partitioning, root dry weight, leaf water content, nutrient concentration, tissue chemistry, root architecture, and final yield components often require harvesting, drying, weighing, washing, chemical analysis, or anatomical inspection. These measurements can be accurate and biologically meaningful, but they interrupt plant growth and prevent continuous observation of the same individual over time. This is a serious limitation for dynamic traits such as stress progression, recovery after irrigation, root zone adaptation, and reproductive development. In root phenotyping, for example, trenching, soil cores, root washing, and shovelomics can reveal belowground traits, but they disturb the root soil system and are difficult to scale across large breeding populations \cite{trachsel2011shovelomics,mcgrail2020trait}. The consequence is that conventional phenotyping often produces accurate snapshots but weak temporal understanding. 

The limitations of manual phenotyping are not simply operational. But they also shape the scientific questions that can be asked. When measurements are slow, expensive, or destructive, researchers tend to select fewer genotypes, fewer time points, fewer environments, or fewer traits. This narrows the ability to detect complex responses involving trait plasticity, genotype by environment interaction, or delayed stress effects. Conventional phenotyping therefore tends to emphasize traits that are easy to measure rather than traits that are most predictive of crop performance. This problem has been highlighted in discussions of plant phenomics, where the field has repeatedly called for non invasive, high throughput, standardized, and data integrated approaches that retain biological relevance while improving scale and resolution \cite{tardieu2017plant,pieruschka2019plant}. In the PhenoAgent framing, these conventional measurements remain essential as ground truth, but they must be connected to continuous sensing and reasoning systems.

\begin{table*}[!t]
\centering
\caption{Conventional phenotyping measurements across the seed soil plant continuum. The table shows why these methods remain valuable biological references, but are inadequate for continuous, scalable, decision-oriented phenotyping.}
\label{tab:conventional_measurements}
\scriptsize
\setlength{\tabcolsep}{3pt}
\renewcommand{\arraystretch}{1.18}

\begin{tabularx}{\textwidth}{@{}
>{\raggedright\arraybackslash}p{2.35cm}
>{\raggedright\arraybackslash}X
>{\raggedright\arraybackslash}X
>{\raggedright\arraybackslash}X
>{\raggedright\arraybackslash}X
>{\raggedright\arraybackslash}p{3.05cm}
@{}}
\toprule
\textbf{Phenotyping layer} 
& \textbf{Typical conventional measurements} 
& \textbf{Common methods} 
& \textbf{Strengths} 
& \textbf{Key limitations} 
& \textbf{References} \\
\midrule

Seed quality and vigor 
& Viability, germination percentage, germination speed, seedling vigor, aging tolerance, stress germination 
& Standard germination test, accelerated aging, electrical conductivity, tetrazolium staining, seedling length and dry weight 
& Directly linked to crop establishment and seed lot value 
& Time consuming, often destructive or semi destructive, limited single seed tracking, weak field prediction under stress 
& Rajjou et al.~\cite{rajjou2012seed}; Marcos Filho~\cite{marcosfilho2015seed}; Reed et al.~\cite{reed2022seed}; Xing et al.~\cite{xing2023seedvigor} \\

Soil/substrate and root zone 
& Moisture, pH, electrical conductivity, salinity, nutrient availability, bulk density, organic matter, microbial or rhizosphere indicators 
& Soil sampling, laboratory chemical analysis, pot/substrate assays, root washing, soil cores, trenching 
& Provides mechanistic context for plant performance 
& Spatially sparse, delayed lab turnaround, destructive sampling, poor linkage to individual plant time series 
& Viscarra Rossel et al.~\cite{viscarra2011proximal}; B{\"u}nemann et al.~\cite{bunemann2018soil}; de la Fuente Cant{\'o} et al.~\cite{delafuente2020extended}; Blanchy et al.~\cite{blanchy2025belowground} \\

Root architecture 
& Root depth, root angle, root length density, root biomass, crown root number, lateral roots, root hair traits 
& Excavation, shovelomics, root washing, rhizoboxes, minirhizotrons, root imaging after harvest 
& Captures belowground traits relevant to water and nutrient uptake 
& Laborious, genotype by soil sensitivity, disturbance of rhizosphere, low throughput in field plots 
& Trachsel et al.~\cite{trachsel2011shovelomics}; Kuijken et al.~\cite{kuijken2015root}; Atkinson et al.~\cite{atkinson2019hidden}; Takahashi and Pradal~\cite{takahashi2021root} \\

Shoot growth and canopy traits 
& Plant height, leaf area, biomass, tillering, canopy cover, chlorophyll, flowering time, plant architecture 
& Rulers, calipers, SPAD meters, visual scoring, leaf area meters, destructive biomass harvest 
& Simple, interpretable, widely used in breeding and agronomy 
& Observer bias, sparse temporal resolution, destructive endpoints, limited spatial coverage 
& Fiorani and Schurr~\cite{fiorani2013future}; Araus and Cairns~\cite{araus2014field}; Watt et al.~\cite{watt2020phenotyping}; Wang et al.~\cite{wang2025sensors} \\

Stress and disease response 
& Wilting, chlorosis, necrosis, disease severity, drought response, salinity response, nutrient deficiency symptoms 
& Visual scales, manual scoring, tissue sampling, physiological assays, yield loss assessment 
& Agronomically meaningful and accepted by breeders 
& Subjective scoring, late symptom visibility, low repeatability among observers, limited early warning ability 
& Poorter et al.~\cite{poorter2016pampered}; Tardieu et al.~\cite{tardieu2017plant}; Murphy et al.~\cite{murphy2024deep} \\

Yield and product quality 
& Grain yield, fruit number, fruit weight, harvest index, quality grades, postharvest traits 
& Manual harvest, combine yield, laboratory quality testing, grading 
& Final economic endpoint and breeding target 
& End of season only, confounded by many earlier processes, weak diagnostic value for cause of failure 
& Ray et al.~\cite{ray2013yield}; Cooper et al.~\cite{cooper2021gem}; Mahmood et al.~\cite{mahmood2022gem} \\

\bottomrule
\end{tabularx}
\end{table*}
\subsection{Seed, Root Zone, and Plant Growth Assessment}

Seed quality and germination represent the first biological stage at which crop performance can diverge. Conventional seed testing evaluates whether a seed lot can germinate under favorable conditions, whether seedlings emerge rapidly and uniformly, and whether vigor is retained after storage or stress exposure. Standard germination tests remain widely used because they are simple and internationally familiar, but germination percentage alone does not fully capture field establishment potential. Seed vigor is a broader concept that includes germination speed, synchrony, stress tolerance, seedling growth, storage performance, and the ability to establish under suboptimal conditions \cite{rajjou2012seed,reed2022seed,jin2025seedphenotyping}. This is why seed scientists have long used complementary tests such as accelerated aging, electrical conductivity, tetrazolium staining, cold tests, seedling length, dry weight, and stress germination assays \cite{marcosfilho2015seed,powell2022seedvigor}. Table \ref{tab:conventional_measurements} demonstrate conventional phenotyping measurements across the seed soil plant continuum.
Recent seed review literature is important for this section because it shows that even seed phenotyping, one of the oldest agricultural assessment tasks, is still constrained by conventional procedures. Reviews on seed vigor describe traditional tests as informative but slow, sample consuming, species dependent, and sometimes weakly predictive of performance under field stress \cite{xing2023seedvigor,li2023opticalseed}. The implication for PhenoAgent is that seed phenotyping should not be treated as a separate pre crop quality control step. Instead, seed level traits should be linked to germination time series, seedling vigor, substrate moisture, salinity, root zone measurements, and later plant growth. This would allow the system to answer a more useful question than whether a seed lot germinated under ideal conditions. It would ask whether early seed and emergence behavior predicts later crop performance under a defined environment and management strategy.
Soil and substrate assessment forms the second layer of the conventional phenotyping continuum. Soil pH, electrical conductivity, salinity, moisture, organic matter, nutrient availability, texture, bulk density, microbial activity, and substrate physical properties all influence root growth and plant response. In greenhouse and controlled environment agriculture, substrates and root zone conditions are especially important because irrigation and fertigation decisions can rapidly change salinity, nutrient concentration, oxygen availability, and moisture distribution. Conventional soil and substrate assessments typically rely on sampling followed by laboratory or handheld measurements. These methods provide important physical and chemical context, but they are often spatially sparse and disconnected from continuous plant observations. Soil quality reviews emphasize that soil health cannot be reduced to a single variable and must be interpreted through physical, chemical, and biological indicators \cite{bunemann2018soil,viscarra2011proximal}.

Root and rhizosphere phenotyping further exposes the limits of conventional assessment. Roots are central to water and nutrient uptake, drought adaptation, salinity response, and crop establishment, yet the root system is difficult to observe because it is hidden inside soil or substrate. Conventional root methods such as excavation, washing, trenching, soil coring, and shovelomics have contributed important knowledge, but they are destructive or semi destructive, laborious, and often unsuitable for repeated measurement of the same plant \cite{trachsel2011shovelomics,kuijken2015root}. Recent root phenotyping reviews argue that root traits and rhizosphere traits must be integrated into crop improvement, but also stress that standardization and field relevance remain major challenges \cite{tracy2020roots,takahashi2021root}. This supports a central argument of the present review, emphasizing that next-generation phenotyping should integrate information from seeds, substrates, root zones, and shoots rather than treating these components as independent datasets. To avoid treating conventional phenotyping as a static historical baseline, Table~\ref{tab:conventional_chronological_studies} summarizes representative studies that progressively exposed the need for higher throughput, less destructive, and more integrated phenotyping. The table shows that the motivation for PhenoAgent does not arise only from recent foundation models, it is rooted in long standing limitations in seed, root zone, field trail, and G$\times$E$\times$M phenotyping.

\begingroup
\scriptsize
\setlength{\tabcolsep}{1.7pt}
\renewcommand{\arraystretch}{1.12}
\setlength{\LTcapwidth}{\linewidth}

\begin{longtable}{@{}
M{0.70cm}
P{2.30cm}
P{1.85cm}
P{2.35cm}
P{2.45cm}
P{2.55cm}
P{2.45cm}
@{}}

\caption{Chronological development of conventional and transitional studies relevant to seed, soil/root zone, plant growth, field trail, and G$\times$E$\times$M phenotyping. The ordering highlights how the field moved from recognizing the phenotyping bottleneck toward integrated but still incomplete seed soil plant assessment.}
\label{tab:conventional_chronological_studies} \\

\toprule
\textbf{Year} & \textbf{Study} & \textbf{Domain} & \textbf{Contribution} & \textbf{Conventional relevance} & \textbf{Limitations} & \textbf{Relevance to PhenoAgent} \\
\midrule
\endfirsthead

\multicolumn{7}{@{}l}{\small\itshape Table~\ref{tab:conventional_chronological_studies} continued from previous page.} \\[2pt]
\toprule
\textbf{Year} & \textbf{Study} & \textbf{Domain} & \textbf{Contribution} & \textbf{Conventional relevance} & \textbf{Limitations} & \textbf{Relevance to PhenoAgent} \\
\midrule
\endhead

\midrule
\multicolumn{7}{r}{\small\itshape Continued on next page.} \\
\endfoot

\bottomrule
\endlastfoot

2011 & Furbank and Tester~\cite{furbank2011phenomics} & Plant phenomics & Framed phenotyping as a bottleneck limiting crop improvement & Established need for higher throughput measurement & Did not yet address modern agentic or multimodal reasoning & Provides historical motivation for automated intelligence \\
2011 & Trachsel et al.~\cite{trachsel2011shovelomics} & Root phenotyping & Introduced shovelomics for field root architecture assessment & Practical method for root crown traits in field breeding & Destructive, labor intensive, limited temporal monitoring & Shows why belowground sensing must become less destructive \\
2012 & Rajjou et al.~\cite{rajjou2012seed} & Seed germination/vigor & Reviewed molecular and physiological bases of germination vigor & Connects seed vigor to establishment success & Molecular markers are not automatically linked to later crop management & Supports seed to plant phenotype continuity \\
2013 & Fiorani and Schurr~\cite{fiorani2013future} & Plant phenotyping & Discussed future scenarios for noninvasive phenotyping & Emphasized resource use traits and environmental response & Large scale data integration remained a challenge & Anticipates the need for integrated phenotyping systems \\
2014 & Araus and Cairns~\cite{araus2014field} & Field phenotyping & Positioned field HTPP as a crop breeding frontier & Linked phenotyping to breeding under realistic environments & Field data remain noisy and difficult to interpret causally & Motivates context aware phenotyping agents \\
2015 & Marcos Filho~\cite{marcosfilho2015seed} & Seed vigor & Reviewed seed vigor testing development and practice & Clarifies why vigor is more informative than germination alone & Many tests remain species dependent and laborious & Supports seed quality reasoning within PhenoAgent \\
2015 & Kuijken et al.~\cite{kuijken2015root} & Root phenotyping & Discussed translation of root traits from lab to breeding & Highlights root component traits for crop improvement & Lab root traits often fail to translate directly to field performance & Shows need to connect root data with environment and management \\
2016 & Poorter et al.~\cite{poorter2016pampered} & Controlled vs field conditions & Compared lab, greenhouse, and field grown plant phenotypes & Demonstrated translation challenges between environments & Controlled conditions may not reproduce field physiology & Supports environment aware interpretation and validation \\
2017 & Tardieu et al.~\cite{tardieu2017plant} & Sensors to knowledge & Argued phenomics must transform sensor data into knowledge & Bridges measurement and biological interpretation & Did not define agentic decision making workflow & Direct conceptual foundation for PhenoAgent \\
2018 & B{\"u}nemann et al.~\cite{bunemann2018soil} & Soil quality & Reviewed physical, chemical, and biological soil quality indicators & Shows soil/substrate condition is multidimensional & Soil health indicators are difficult to connect to individual plant trajectories & Supports root zone layer in seed soil plant intelligence \\
2020 & Watt et al.~\cite{watt2020phenotyping} & Breeding phenotyping & Reviewed new phenotyping windows for breeders & Covered seed, shoot, root, field, and controlled phenotyping & Integration and FAIR data remain challenging & Supports breeder-facing PhenoAgent framework \\
2020 & Tracy et al.~\cite{tracy2020roots} & Root phenotyping & Highlighted expanding opportunities in root phenotyping for crop improvement & Shows breeding value of root traits and rhizosphere selection & Root phenotyping remains slow and difficult under real soil & Supports belowground phenotyping and digital root zone records \\
2020 & McGrail et al.~\cite{mcgrail2020trait} & Root trait selection & Argued trait-based root phenotyping is needed for crop selection & Connects root traits to soil type and climate scenarios & Root traits are often ignored in breeding pipelines & Justifies soil-root integration in PhenoAgent \\
2021 & Cooper et al.~\cite{cooper2021gem} & G$\times$E$\times$M prediction & Proposed pathways to close yield gaps through G$\times$E$\times$M prediction & Shows management must be considered with genotype and environment & Full factorial exploration is difficult empirically & Motivates agentic reasoning over genotype, environment, and management \\
2021 & Takahashi and Pradal~\cite{takahashi2021root} & Root modeling & Defined minimum root information needed for root modeling & Helps reuse legacy root phenotyping data & Root trait standardization remains incomplete & Supports structured root digital twin representation \\
2022 & Reed et al.~\cite{reed2022seed} & Seed vigor under climate stress & Linked germination vigor to crop sustainability under climate change & Shows seed vigor affects establishment under variable conditions & Seed quality often remains disconnected from later crop analytics & Supports early-stage prediction in PhenoAgent \\
2022 & Mahmood et al.~\cite{mahmood2022gem} & GEM reciprocity & Reviewed genotype-environment-management reciprocity and productivity & Highlights management as a co-determinant of phenotype & Limited operational integration with phenotyping data streams & Supports (SSPEM) continuum \\
2023 & Xing et al.~\cite{xing2023seedvigor} & Seed vigor testing & Reviewed physiological changes and nondestructive test methods & Summarized traditional and emerging seed-vigor methods & Calls for faster, nondestructive, mechanistically linked tools & Supports multimodal seed phenotyping \\
2023 & Li et al.~\cite{li2023opticalseed} & Optical seed sensing & Reviewed optical sensing for seed vigor determination & Shows transition from manual seed tests to rapid sensing & Seed sensing often remains isolated from crop development & Supports integration of seed sensing with crop digital records \\
2024 & Murphy et al.~\cite{murphy2024deep} & Image based AI phenotyping & Reviewed deep learning in image based phenotyping & Shows AI can reduce manual image interpretation & AI models still depend on labels and task-specific datasets & Supports transition from manual scoring to automated perception \\
2025 & Blanchy et al.~\cite{blanchy2025belowground} & Belowground field phenotyping & Demonstrated non invasive belowground phenotyping infrastructure & Addresses the weak belowground layer of field phenotyping & Requires expert processing and integration with aboveground data & Supports whole plant and root zone data fusion \\
2025 & Wang et al.~\cite{wang2025sensors} & Image based HTPP trends & Reviewed sensors, platforms, deep learning, and digital trends & Summarizes technological progression beyond conventional methods & Agentic decision making remains limited & Provides bridge from conventional to PhenoAgent systems \\

\end{longtable}
\endgroup
\subsection{Greenhouse, Field, and Breeding Trial Phenotyping}

Field trials and greenhouse trials are the experimental backbone of crop improvement. Field trials evaluate genotype performance under realistic production conditions, including natural variability in weather, soil, pests, disease pressure, and management. Greenhouse and growth chamber trials provide stronger control over temperature, humidity, light, irrigation, nutrient supply, and stress treatments. Breeding pipelines use both trial types because controlled environments can accelerate early screening, while field trials remain essential for evaluating yield, adaptation, stability, and commercial performance. However, translating phenotypes from controlled conditions to field performance is not straightforward. A major review comparing plants grown inside and outside showed that controlled environment plants can differ substantially from field grown plants in growth rate, morphology, physiology, plant density response, and environmental exposure, with only modest correspondence between lab and field phenotypic data \cite{poorter2016pampered,watt2020phenotyping}.
In conventional breeding pipelines, phenotyping is usually organized as a staged selection process. Early generations may be screened for germination, vigor, disease resistance, plant architecture, flowering time, or stress tolerance using controlled or semi controlled methods. Later generations are tested across multiple locations and seasons to evaluate yield, stability, and adaptation. This staged design is practical, but it also introduces discontinuities. Seed data may not be linked to later plant data, greenhouse stress assays may not capture field complexity, soil measurements may be taken at plot level rather than plant level and final yield may be analyzed without a detailed causal record of the developmental events that produced it. The field high throughput phenotyping literature has argued that better sensing is needed in breeding because conventional field scoring is too slow and sparse to capture the temporal and spatial dimensions of crop performance \cite{araus2014field,tardieu2017plant}.
Greenhouse and controlled environment agriculture create a special case for conventional phenotyping. Unlike open field trials, management variables such as irrigation, fertigation, lighting, temperature, humidity, CO2, and airflow can be controlled or modified frequently. This makes greenhouse trials powerful for studying cause and effect relationships, but only if phenotyping can capture the response with sufficient detail. Conventional greenhouse phenotyping often relies on periodic manual measurements, visual scoring, end point harvest, and separate sensor logs. This creates the same fragmentation problem seen in field trials. The plant is measured in one record, the environment in another, the management events in another, and the final yield in another. For PhenoAgent, the opportunity is to convert greenhouse trials into continuous learning experiments in which seed, substrate, plant, environment, and management are linked through time.

\subsection{Genotype Environment Management Interactions in Phenotyping}

A phenotype is not produced by genotype alone. It emerges from interactions among genotype, environment, and management, often described as G$\times$E or G$\times$E$\times$M interactions. In practical terms, the same genotype may perform differently across environments, and the same environment may produce different outcomes depending on irrigation, planting density, fertilizer regime, substrate, sowing date, or climate control strategy. This is why multi environment trials are central to breeding and why crop management cannot be separated from phenotyping. Recent reviews of G$\times$E and G$\times$E$\times$M interactions emphasize that crop improvement requires identifying genotypes and management practices that perform reliably across target environments or exploit specific niches \cite{de2021gxe,cooper2021gem}. For arid region and controlled environment agriculture, this interaction space is even more important because water, salinity, heat, and substrate conditions strongly affect performance.
Conventional phenotyping struggles with G$\times$E$\times$M interactions because the factorial space is too large to explore manually. A trial that compares several genotypes, seed lots, substrates, irrigation regimes, salinity levels, nutrient formulations, and environmental settings quickly becomes expensive and operationally complex. As a result, many studies simplify the system by testing fewer factors, fewer time points, or fewer traits. This simplification can hide important causal mechanisms. For example, a genotype may appear low yielding under one substrate not because its genetic potential is poor, but because its root architecture interacts poorly with moisture distribution or salinity accumulation. Similarly, a seed lot may appear acceptable in a standard germination test but fail under greenhouse stress because vigor, substrate, and irrigation timing interact unfavorably. G$\times$E$\times$M literature therefore strengthens the argument that phenotyping must evolve from trait measurement toward systems level reasoning \cite{mahmood2022gem,cooper2021gem}.
\begin{figure}[h!]
    \centering
    \includegraphics[width=0.8\textwidth]{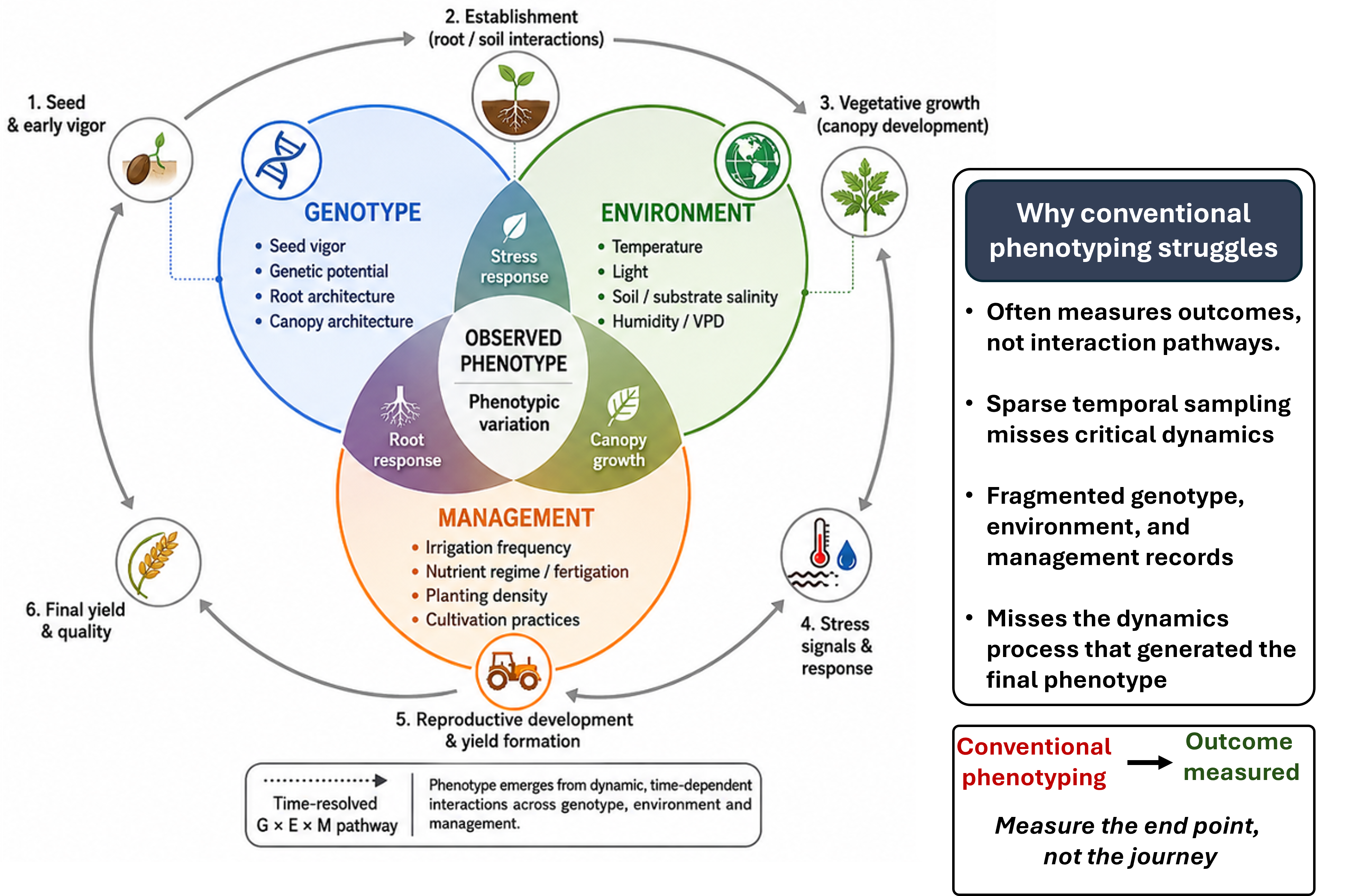}
    \caption{Dynamic genotype environment management (GEM) interactions shaping plant phenotypes over time. Genetic potential, environmental exposure, and management practices jointly influence root development, canopy growth, stress response, yield formation, and quality. The pathway highlights the progression from seed vigor and establishment to vegetative growth, stress adaptation, reproductive development, and final crop outcome, emphasizing the need for time-resolved phenotyping to interpret measured traits.}

    \label{fig:gem_interactions_conventional_phenotyping}
\end{figure}
The G$\times$E$\times$M perspective also clarifies why PhenoAgent should not be framed merely as an automated dashboard. The scientific challenge is to build a phenotyping system that can track interacting causes over time and use them to explain observed plant states. Conventional phenotyping can identify that a crop is shorter, lower yielding, chlorotic, or delayed in flowering, but it often cannot reconstruct the causal chain linking seed quality, substrate condition, irrigation history, environmental exposure, and genotype response. A PhenoAgent style system would treat this causal chain as the central object of phenotyping. It would not replace conventional trials, rather, it would enrich them by transforming trial plots, greenhouse beds, or plant trays into temporally resolved biological experiments. Figure \ref{fig:gem_interactions_conventional_phenotyping} summarizes genotype environment management interaction in phenotyping. 
\subsection{Limitations of Conventional Phenotyping}

Limitations of conventional phenotyping include labor, time, subjectivity, limited scale, and delayed decisions. Labor is most apparent as manual measurements require trained staff, repeated visits, strict protocols, and substantial time, limiting how many plants, plots, seed lots, substrates, or treatments can be assessed. Time is also critical because measurements are often taken infrequently or only at harvest, capturing outcomes but missing dynamics such as stress onset and recovery, growth changes, or developmental transitions. Subjectivity affects visual scoring of disease, chlorosis, wilting, lodging, and vigor, where observer experience, lighting, fatigue, and scoring scales reduce consistency. These issues have driven imaging and deep learning, yet conventional scores still serve as reference labels for many AI systems \cite{minervini2015image,murphy2024deep}.
Scale creates a second-order problem as sparse phenotyping leads researchers to interpret end traits without intermediate context. A yield drop may be blamed on genotype when drivers include seed vigor, soil variability, irrigation, disease timing, or environmental stress. A greenhouse treatment may seem effective, yet manual measurements may miss whether gains came from earlier emergence, stronger root zones, faster canopy closure, or reduced late stress. Hence data integration is central in phenomics. Phenotyping data are valuable not only for accuracy but for linkages across traits, time, environments, and experiments \cite{coppens2017unlocking,papoutsoglou2020miappe}. Conventional phenotyping provides ground truth but seldom a complete digital history of crop development.
Delayed decision making most directly limits agricultural impact. In breeding, slow phenotyping delays selection and raises multi-year trial costs. In greenhouses, late detection of poor germination, root-zone imbalance, salinity, nutrient stress, or disease reduces uniformity and yield. In seed and substrate optimization, delayed feedback obscures which early conditions drove later performance. Although belowground and high-throughput phenotyping is moving toward continuous, less invasive measurement, many systems still focus on data acquisition and analysis rather than actionable reasoning \cite{blanchy2025belowground,wang2025sensors}. This gap motivates the shift from conventional phenotyping to current and next-generation approaches. Table \ref{tab:conventional_limitations_phenoagent} summarizes conventional limitations and their implications for PhenoAgent-based next-generation phenotyping.
\begin{table}[h!]
\centering
\caption{Limitations of conventional phenotyping and their implications for PhenoAgent based next generation phenotyping.}
\label{tab:conventional_limitations_phenoagent}
\scriptsize
\setlength{\tabcolsep}{2.2pt}
\renewcommand{\arraystretch}{1.15}

\begin{tabularx}{\linewidth}{@{}
>{\raggedright\arraybackslash}p{2.0cm}
>{\raggedright\arraybackslash}X
>{\raggedright\arraybackslash}X
>{\raggedright\arraybackslash}X
>{\raggedright\arraybackslash}p{2.45cm}
@{}}
\toprule
\textbf{Limitation} 
& \textbf{How it appears in conventional phenotyping} 
& \textbf{Scientific consequence} 
& \textbf{Operational consequence} 
& \textbf{PhenoAgent opportunity} \\
\midrule

Labor intensity 
& Manual scoring, plant measurement, seedling counting, root washing, destructive harvest, soil sampling 
& Limits population size, trait diversity, and temporal frequency 
& Increases cost and reduces commercial scalability 
& Automate observation using sensors, robots, and AI assisted annotation \\

Destructive endpoints 
& Biomass harvest, root excavation, tissue analysis, seedling dissection 
& Prevents continuous tracking of the same plant or plot 
& Delays understanding of stress progression and recovery 
& Link non-destructive sensing with selective ground-truth sampling \\

Subjectivity 
& Visual vigor, disease severity, wilting, lodging, chlorosis and canopy scores 
& Reduces reproducibility and introduces observer bias 
& Limits trust and comparability across sites and operators 
& Use calibrated models with uncertainty estimates and expert review \\

Sparse temporal data 
& Measurements collected weekly, at key stages, or only at harvest 
& Misses transient stress, early divergence, and treatment-response timing 
& Prevents timely intervention in greenhouse and field trials 
& Build temporal digital plant records and growth trajectories \\

Separated data streams 
& Seed, soil, weather, irrigation, image, and yield data stored independently 
& Weak causal interpretation across the seed soil plant continuum 
& Makes trial decisions reactive rather than predictive 
& Fuse seed, soil/root zone, plant, environment, and management data \\

Weak G$\times$E$\times$M coverage 
& Limited factorial combinations and few environments 
& Important interactions remain hidden or confounded 
& Slows breeding and management optimization 
& Use agentic reasoning to prioritize measurements, treatments, and interventions \\

Delayed decisions 
& Stress and poor establishment detected after visible symptoms or final yield loss 
& Reduces ability to learn cause-and-effect early 
& Lost yield, wasted inputs, and slower trial cycles 
& Generate early warnings, explanations, and intervention recommendations \\

\bottomrule
\end{tabularx}
\end{table}

Conventional phenotyping remains indispensable because it provides biological definitions, reference measurements, and validation data. However, its limitations are structural rather than incidental. Manual and destructive methods cannot easily capture continuous, multiscale, and multimodal crop development across the SSPEM continuum. Field and greenhouse trials remain essential, but they need richer temporal records that connect early seed and substrate conditions to later plant performance. G$\times$E$\times$M interactions explain why crop outcomes cannot be understood from isolated traits alone. These observations establish the bridge to the next section, current advances in high throughput sensing, robotics, IoT, deep learning, digital twins, and multimodal data fusion are attempts to overcome the conventional phenotyping bottleneck, but they must ultimately be integrated into reasoning systems such as PhenoAgent.

\section{Current Advances in Automated and AI-Driven Phenotyping}
\label{sec-3}

\begin{figure}[t!]
    \centering
    \includegraphics[width=1\textwidth]{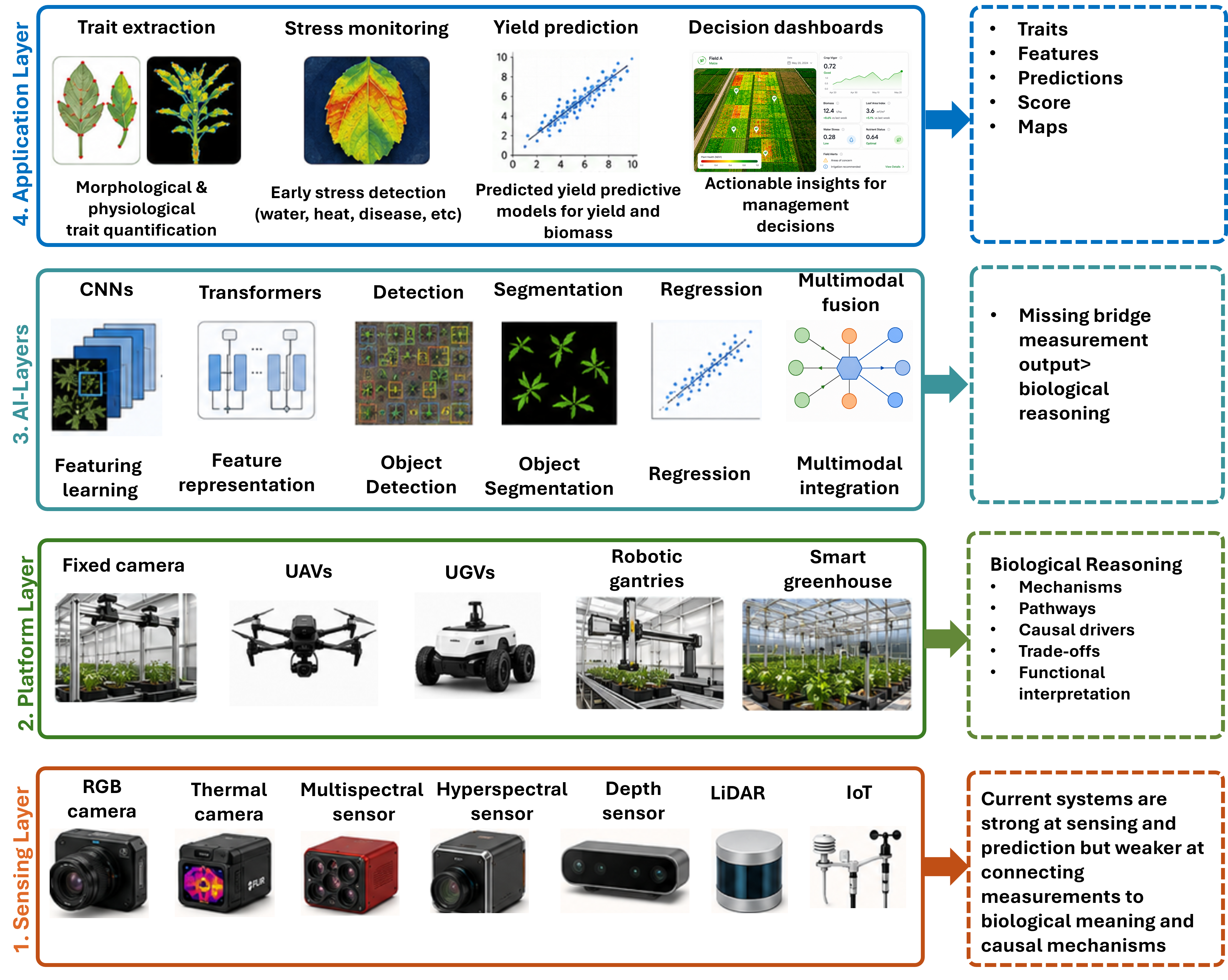}
    \caption{Current high-throughput phenotyping ecosystem. The framework summarizes the main sensing modalities, data-acquisition platforms, AI methods, and application outputs used for automated crop monitoring, including trait extraction, stress assessment, yield prediction, and decision dashboards. It also highlights the remaining gap between measurement-driven outputs and biologically grounded crop-state reasoning.}
    \label{fig:current_high_throughput_phenotyping_ecosystem}
\end{figure}

Current plant phenotyping is moving from slow human observation toward high throughput sensing, robotic data acquisition, and AI assisted interpretation. This transition has been driven by two simultaneous changes. First, the rapid availability of low cost RGB cameras, thermal cameras, multispectral and hyperspectral sensors, depth cameras, LiDAR units, IoT devices, and edge-computing hardware. Second, the emergence of deep learning models that can detect organs, segment plants, classify stress, extract traits, predict yield, and fuse heterogeneous data streams \cite{furbank2011phenomics,tardieu2017plant,murphy2024deep}. These advances have made it possible to measure crop development at scales and frequencies that were previously unrealistic. Fixed cameras can monitor the same plant every hour, UAVs can scan field trials across entire breeding blocks, UGVs can collect close-range canopy and root zone observations, and smart greenhouses can synchronize image data with climate, irrigation, fertigation, and actuator logs \cite{busemeyer2013breedvision,atefi2021robotic,xie2021uavmultisensor}.
This section surveys recent progress in sensing technologies, platforms, deep learning, foundation models, multimodal fusion, and digital twins, and then highlights the gap between stand-alone high-throughput phenotyping and an integrated understanding of seed–soil–plant–environment management. 
Figure~\ref{fig:current_high_throughput_phenotyping_ecosystem} illustrates the current high-throughput phenotyping pipeline, from sensing and robotic data acquisition to AI-based trait extraction, stress monitoring, yield prediction, and dashboard visualization. While these systems improve measurement speed and scale, they often remain focused on outputs rather than biological interpretation, leaving a gap between automated observation and explainable crop-state reasoning.

\subsection{Imaging and Sensing Technologies}
The sensor layer is the foundation of modern phenotyping because each sensing modality captures a different expression of plant status. RGB imaging remains the most widely used modality because it is inexpensive, high resolution, easy to deploy, and suitable for visible traits such as canopy cover, leaf area, color, plant architecture, flower count, fruit count, disease lesions, lodging, and emergence. The rise of computer vision in plant phenotyping was enabled by RGB image repositories, open source image analysis pipelines, and deep learning methods that reduced the need for hand crafted features \cite{minervini2015image,pound2017deep,ubbens2017deep}. However, RGB alone is often limited when stress symptoms are subtle, when leaves overlap, when lighting varies, or when physiological status changes before visible symptoms appear. This limitation has pushed phenotyping toward multimodal sensing rather than single image analysis.

Thermal imaging provides canopy temperature as a proxy for transpiration, stomatal conductance, water stress, and energy balance. It is particularly useful in drought, salinity, irrigation scheduling, and greenhouse climate studies, but thermal data are sensitive to viewing geometry, wind, humidity, radiation, calibration, and temporal conditions \cite{perich2020canopy,xie2021uavmultisensor,upadhyay2025plantdiseasecv}. Multispectral and hyperspectral imaging extend phenotyping from morphology toward biochemical and physiological assessment by measuring reflectance across selected bands or dense spectral ranges. These modalities support vegetation indices, pigment estimation, nitrogen status, water content, disease detection, and stress monitoring, yet they require calibration, spectral preprocessing, and careful interpretation because spectral responses can be confounded by canopy structure, soil background, growth stage, and illumination \cite{sankaran2015uav,nguyen2023uavmultisensory,wang2025sensors}.

Depth cameras and LiDAR add structural information that is difficult to recover from two dimensional imagery. They can estimate plant height, canopy volume, lodging, organ geometry, row structure, biomass proxies, and three dimensional canopy architecture. When combined with hyperspectral or thermal data, structural sensing can help separate biochemical signals from canopy size effects and reduce saturation in dense vegetation \cite{xie2021uavmultisensor,nguyen2023uavmultisensory}. IoT sensors complement imaging by continuously recording environmental and management context, including temperature, humidity, light intensity, CO2, substrate moisture, pH, electrical conductivity, nutrient solution properties, irrigation events, and actuator status. In controlled environment agriculture, this contextual layer is essential because crop phenotype is tightly shaped by microclimate and fertigation decisions \cite{wolfert2017big,walter2017smart,rahman2024digitaltwin}. Table \ref{tab:current_sensing_modalities} summarizes major sensing modalities used current phenotyping.
\begin{table*}[h!]
\centering
\caption{Key sensing modalities employed in today’s high-throughput phenotyping and how they relate to seed-soil-plant intelligence. The table highlights that while each modality provides useful information, none is sufficient on its own when considered in isolation.}
\label{tab:current_sensing_modalities}
\scriptsize
\setlength{\tabcolsep}{2.2pt}
\renewcommand{\arraystretch}{1.15}
\resizebox{\textwidth}{!}{%
\begin{tabular}{P{2.3cm} P{3.1cm} P{3.0cm} P{3.0cm} P{3.0cm} P{3.1cm}}
\toprule
\textbf{Modality} & \textbf{Traits / Signals} & \textbf{Deployment} & \textbf{Strength} & \textbf{Current limitation} & \textbf{References} \\
\midrule
RGB imaging & Emergence, leaf area, canopy cover, color, organ count, lesions, fruit count, growth rate & Fixed cameras, UGVs, UAVs, smartphones, greenhouse stations & Low cost, high spatial detail, compatible with deep learning & Sensitive to illumination, occlusion, background, and visible-symptom delay & Minervini et al.~\cite{minervini2015image}; Pound et al.~\cite{pound2017deep}; Tausen et al.~\cite{tausen2020greenotyper}; Meraj et al.~\cite{meraj2024survey} \\
Thermal imaging & Canopy temperature, water stress, transpiration proxy, irrigation response & UAVs, fixed greenhouse cameras, gantry systems & Enables early water-status monitoring beyond RGB appearance & Requires calibration and correction for weather, angle, and emissivity & Perich et al.~\cite{perich2020canopy}; Xie and Li~\cite{xie2021uavmultisensor}; Nguyen et al.~\cite{nguyen2023uavmultisensory} \\
Multispectral imaging & Vegetation indices, chlorophyll, biomass proxy, nitrogen, stress indices & UAVs, handheld sensors, greenhouse cameras & Balances cost and physiological sensitivity & Band selection and vegetation-index saturation can limit generality & Sankaran et al.~\cite{sankaran2015uav}; Xie and Li~\cite{xie2021uavmultisensor}; Khuimphukhieo and Silva~\cite{khuimphukhieo2025uas} \\
Hyperspectral imaging & Pigment, water, nitrogen, disease, salinity and biochemical fingerprints & UAVs, proximal carts, laboratory and greenhouse imaging & Dense spectral information for subtle stress and composition changes & High data volume, calibration burden, limited interpretability without models & Nguyen et al.~\cite{nguyen2023uavmultisensory}; de Silva and Brown~\cite{desilva2023multispectral}; Wang et al.~\cite{wang2025sensors} \\
Depth and LiDAR & Plant height, canopy volume, row structure, lodging, 3D architecture, biomass proxy & UAV LiDAR, UGV LiDAR, gantry systems, RGB-D cameras & Adds spatial structure and geometry, reducing ambiguity of 2D images & Cost, registration, occlusion, point-cloud processing, and cross-platform calibration & Busemeyer et al.~\cite{busemeyer2013breedvision}; Xie and Li~\cite{xie2021uavmultisensor}; Nguyen et al.~\cite{nguyen2023uavmultisensory} \\
IoT and environmental sensors & Temperature, humidity, light, CO2, substrate moisture, pH, EC, irrigation, fertigation, actuator logs & Smart greenhouses, vertical farms, field sensor networks & Provides continuous causal context for plant response & Often stored separately from image-derived traits and biological labels & Wolfert et al.~\cite{wolfert2017big}; Peladarinos et al.~\cite{peladarinos2023digital}; Rahman et al.~\cite{rahman2024digitaltwin} \\
\bottomrule
\end{tabular}}
\end{table*}
\subsection{Automated Phenotyping Platforms}
Automated platforms determine the spatial and temporal scale at which phenotyping can be performed. Fixed camera systems are particularly useful in greenhouses, growth chambers, germination rooms, and vertical farms because they provide repeated observation of the same plants under controlled viewpoints. Distributed camera systems such as greenotyper demonstrated that low cost networked cameras can monitor thousands of plants and support deep learning based segmentation at large image volumes \cite{tausen2020greenotyper}. Fixed platforms are well suited for time series growth monitoring, emergence tracking, leaf expansion, stress progression, and treatment response experiments. However fixed cameras can observe only selected viewpoints unless the facility is densely instrumented or cameras are mounted on rails.

UAV based phenotyping has become one of the strongest trends in field high throughput phenotyping because UAVs can repeatedly scan breeding trials with RGB, multispectral, thermal, hyperspectral, and LiDAR sensors. UAVs are valuable for plot level canopy cover, plant height, vegetation indices, canopy temperature, biomass prediction, flowering, lodging, and yield estimation \cite{sankaran2015uav,maes2019uav,khuimphukhieo2025uas}. Their strength is rapid coverage over large field areas and the limitation is to observe crops mostly from above, which can miss under canopy traits, root zone conditions, plant level identity, and close range organ details. UGVs and robotic carts address this limitation by collecting closer range imagery from side views, under canopy views, and repeatable trajectories. Robotic phenotyping reviews emphasize that UGVs can combine perception, navigation, sample localization, and repeated measurements, but field robustness, localization, battery life, data synchronization, and crop contact safety remain practical barriers \cite{atefi2021robotic,zhang2026ground}.

\begin{figure}[h!]
    \centering
    \includegraphics[width=1\textwidth]{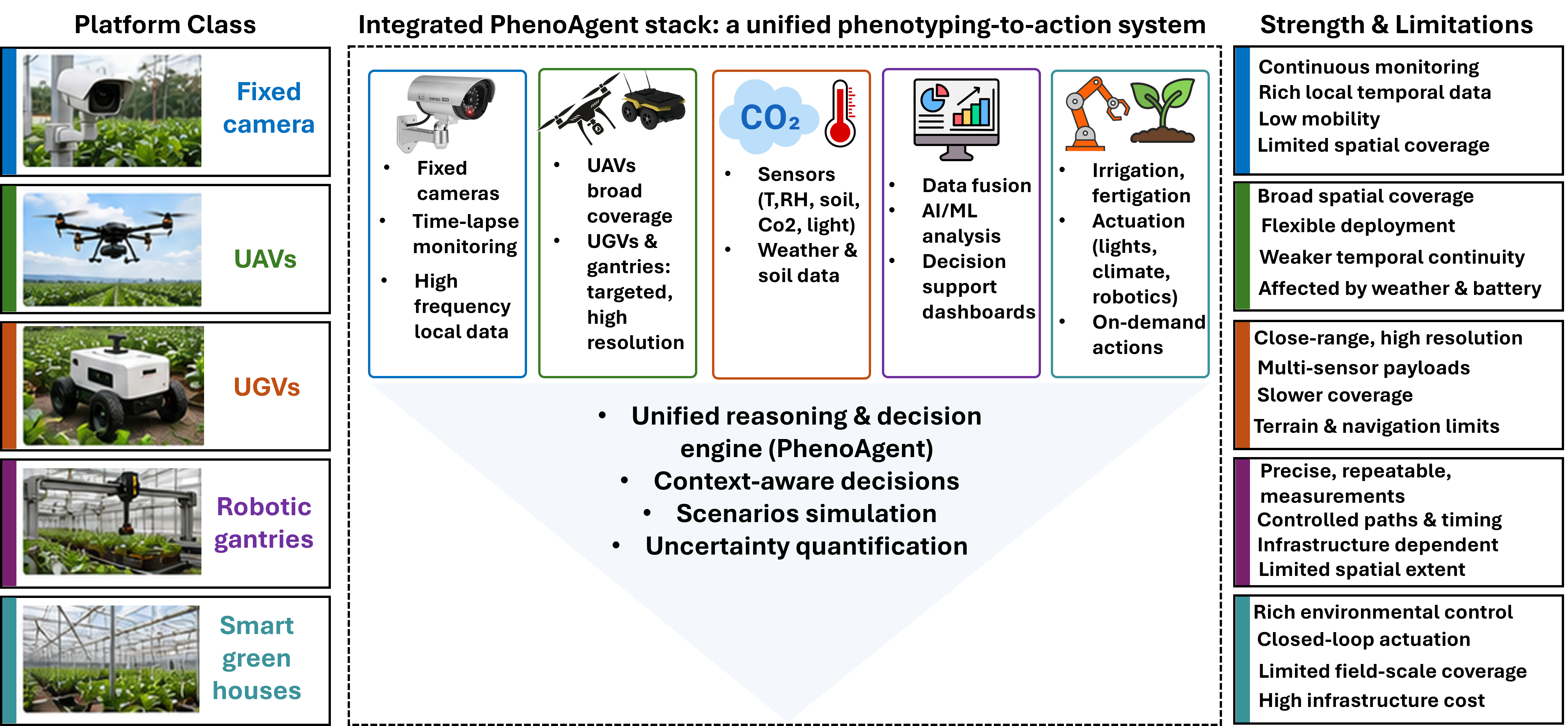}
    \caption{Platform-level comparison of fixed cameras, UAVs, UGVs, robotic gantries, and smart greenhouses. The comparison highlights their complementary strengths and limitations in spatial coverage, temporal frequency, sensing resolution, mobility, environmental context, and potential for closed-loop crop management.}
    \label{fig:platform_comparison_fixed_uav_ugv_greenhouse}
\end{figure}

Smart greenhouses integrate imaging platforms, IoT sensors, climate control systems, fertigation systems, and decision dashboards. This makes them especially relevant to PhenoAgent because the system can observe not only the plant phenotype but also the intervention history that produced it. Digital twin greenhouse studies show that sensing, machine learning, and simulation can be coupled to predict climate, crop response, disease risk, energy use, and actuator settings \cite{slob2022digital,purcell2023digital,rahman2024digitaltwin}. However, many greenhouse platforms still focus on monitoring and control separately. For instance, the sensor dashboard reports the environment, the image model reports plant status, and the grower must manually infer biological cause and management action. 
Figure~\ref{fig:platform_comparison_fixed_uav_ugv_greenhouse} compares major phenotyping platforms, including fixed cameras, UAVs, UGVs, robotic gantries, and smart greenhouses. Each platform offers different strengths in coverage, frequency, resolution, mobility, and environmental context, showing why integrated systems are needed for robust crop monitoring.

\subsection{Deep Learning for Trait Extraction and Prediction}
Deep learning has become the dominant analytical engine for image based phenotyping because it learns visual features directly from data and can handle complex plant shapes, overlapping leaves, variable backgrounds, and high image volumes. Early deep phenotyping studies showed that convolutional neural networks could localize plant organs, classify root and shoot features, and automate tasks that were previously dependent on hand crafted image processing rules \cite{pound2017deep,ubbens2017deep,jin2025seedphenotyping,lei2024agriimagesegmentation}. Subsequent surveys of deep learning in agriculture and plant phenotyping documented the expansion of CNNs, object detectors, semantic segmentation networks, recurrent models, and regression models across disease detection, organ counting, biomass estimation, fruit detection, weed mapping, stress classification, and yield prediction \cite{kamilaris2018deep,liakos2018machine,jiang2020deep,murphy2024deep}.

Detection and segmentation are especially important because many phenotyping traits are derived from correctly separating plants or organs from the background. U-Net like architectures, Mask R-CNN, YOLO variants, DeepLab style networks, and lightweight detectors have been used to segment leaves, detect fruits, identify wheat heads, count organs, and monitor field plants. Studies such as the Global Wheat Head Detection dataset helped move plant phenotyping toward public benchmarks and model comparison, while YOLO based tomato trait and field plant detection studies illustrate the practical shift toward faster, deployable models \cite{david2020global,cardellicchio2023yolo,li2024dcyolo}. However, segmentation accuracy alone does not guarantee biological usefulness. A segmented leaf must be converted into a trait, linked to growth stage, interpreted under environmental context, and validated against agronomic outcomes.

Deep learning has also become central to stress monitoring and yield prediction. Plant disease detection studies demonstrated that CNNs can classify leaf symptoms and outperform manual visual screening under controlled image conditions \cite{mohanty2016disease,ferentinos2018deep,too2019comparative,pacal2024plantdisease}. Broader disease review and model comparison studies further show that high accuracy is possible across benchmark datasets, but that explainability, dataset bias, and field transfer remain persistent concerns \cite{saleem2019plant,thakur2023vitcnn}. Yield prediction studies showed that machine learning and deep neural networks can learn relationships between environmental, management, genetic, and remote sensing features \cite{khaki2019yield,nguyen2023uavmultisensory}. Yet current models often remain crop specific, dataset specific, sensor specific, and stage specific. Many models are trained using labels that are themselves derived from manual scoring, and performance can drop when models are transferred across greenhouse facilities, field sites, growth stages, genotypes, lighting conditions, or imaging devices. This is why the next step is not simply deeper networks, but robust, explainable, multimodal systems that know when model confidence is insufficient and when additional sensing or expert validation is needed.

\subsection{Foundation Models for Agricultural Vision}

Foundation models and transformer architectures are reshaping agricultural vision by shifting the focus from training a separate model for every narrow task toward reusable representations, promptable segmentation, and multimodal reasoning. Vision transformers introduced attention based image representation learning and showed that global context can be modeled through image patches rather than only local convolutional filters \cite{dosovitskiy2021vit}. Agricultural studies have adapted ViTs and hybrid CNN-transformer models for plant disease classification, multispectral disease recognition, and lightweight mobile diagnosis \cite{borhani2022vit,parez2023greenvit,li2023pmvt,gole2023trincnet}. These models are attractive because plant phenotyping often requires attention to both local features, such as lesions or leaf edges, and global context, such as canopy structure and growth stage.

Promptable models such as the Segment Anything Model have created a new opportunity for agricultural image annotation and rapid trait extraction \cite{kirillov2023segment}. Vision language pretraining also showed that image representations can be aligned with natural language supervision, creating a path toward models that can connect visual crop evidence with textual crop knowledge, protocols, and expert descriptions \cite{radford2021clip}. Instead of training a segmentation network from scratch for every crop and imaging setup, users can provide prompts, boxes, points, or masks to segment objects more flexibly. This is important for phenotyping because annotation is one of the largest barriers to model development. However, general vision foundation models are not automatically crop aware. They may segment visible objects but fail to understand plant organs, genotype specific morphology, disease symptom progression, growth stage, or root zone relationships. Reviews of foundation models in smart agriculture therefore emphasize the need for agricultural pretraining, domain adaptation, multimodal alignment, and validation under real production variability \cite{li2024foundation,bommasani2021foundation}.

The significance of foundation models for PhenoAgent is not only better perception. Their larger value is that they can become reusable perception modules inside a reasoning framework. A PhenoAgent could use a promptable segmentation model to isolate leaves, a ViT to classify stress patterns, a time series model to track growth rate, a retrieval system to compare historical trials, and a language model to generate an agronomic explanation with uncertainty and suggested next measurements. Current literature has begun to explore foundation and promptable models, but most systems still stop at visual task performance rather than integrating perception with crop physiology, management history, and intervention planning.

\subsection{Multimodal Fusion for Crop State Modeling}

Multimodal fusion is the key step that converts high throughput phenotyping from image analysis into crop intelligence. At the sensor level, fusion can combine RGB morphology, hyperspectral reflectance, thermal temperature, LiDAR structure, and IoT environmental context. At the biological level, fusion should connect seed vigor, germination, substrate conditions, root zone dynamics, canopy growth, stress symptoms, management events, and yield. UAV multisensory studies show that combining hyperspectral, thermal, and LiDAR data can improve prediction of maize traits and reduce limitations of single modality measurements, especially vegetation index saturation and missing structural context \cite{nguyen2023uavmultisensory,xie2021uavmultisensor}. However, most fusion approaches remain technical combinations of features rather than mechanistic representations of crop development.

Fusion strategies can be grouped into early fusion, intermediate fusion, late fusion, and decision level fusion. Early fusion concatenates raw or preprocessed sensor channels but can be sensitive to registration errors and missing modalities. Intermediate fusion learns shared representations from separate modality encoders and is increasingly compatible with transformer architectures. Late fusion combines predictions from separate models and is easier to implement when modalities have different sampling frequencies. Decision level fusion integrates model outputs with expert rules, crop models, and management constraints. For PhenoAgent, the most important direction is not one fusion architecture alone, but a layered fusion strategy in which data are aligned by plant identity, plot, treatment, growth stage, timestamp, and management event \cite{coppens2017unlocking,papoutsoglou2020miappe,wolfert2017big}.

Current fusion literature also exposes a major gap as seed and soil/substrate information are rarely treated as first class phenotyping modalities in AI pipelines. Most high throughput systems focus on aboveground canopy traits, while seed quality, germination timing, substrate moisture, salinity, EC, nutrient profile, and irrigation/fertigation history are stored in separate agronomic records. This separation limits causal interpretation. For example, a canopy model may detect reduced growth, but without seed lot vigor, substrate EC, moisture distribution, and irrigation history, it cannot distinguish poor emergence from salinity stress, nutrient imbalance, genetic susceptibility, or environmental shock. Therefore, the novel contribution of a PhenoAgent review is to argue that multimodal fusion must be organized around the SSPEM continuum rather than around sensor convenience.

\begingroup
\scriptsize
\setlength{\tabcolsep}{1.7pt}
\renewcommand{\arraystretch}{1.12}
\setlength{\LTcapwidth}{\linewidth}

\begin{longtable}{@{}
M{0.75cm}
P{2.55cm}
P{2.1cm}
P{2.9cm}
P{3.05cm}
P{3.0cm}
P{3.0cm}
@{}}

\caption{Chronological map of representative advances in high throughput, robotic, AI driven, multimodal, and digital twin phenotyping. The studies are ordered by year to show how the field evolved from platform construction and image analysis toward deep learning, fusion, foundation models, and decision oriented digital systems.}
\label{tab:current_chronological_advances} \\

\toprule
\textbf{Year} & \textbf{Study} & \textbf{Domain} & \textbf{Main advance} & \textbf{Contribution} & \textbf{Remaining gap} & \textbf{Relevance to PhenoAgent} \\
\midrule
\endfirsthead

\multicolumn{7}{@{}l}{\small\itshape Table~\ref{tab:current_chronological_advances} continued from previous page.} \\[2pt]
\toprule
\textbf{Year} & \textbf{Study} & \textbf{Domain} & \textbf{Main advance} & \textbf{Contribution} & \textbf{Remaining gap} & \textbf{Relevance to PhenoAgent} \\
\midrule
\endhead

\midrule
\multicolumn{7}{r}{\small\itshape Continued on next page.} \\
\endfoot

\bottomrule
\endlastfoot

2011 & Furbank and Tester~\cite{furbank2011phenomics} & Phenomics vision & Defined phenotyping bottleneck & Motivated sensor-based scale-up & Limited AI and robotics integration & Historical baseline for automation \\
2012 & White et al.~\cite{white2012field} & Field phenomics & Framed field-based phenomics for genetics & Linked field measurements to genetic discovery & Sparse automation compared with current systems & Supports field-relevant validation \\
2013 & Busemeyer et al.~\cite{busemeyer2013breedvision} & Multi-sensor platform & Tractor-based BreedVision system & Combined sensors for field breeding & Heavy platform, limited reasoning layer & Demonstrates platform-level integration \\
2013 & Cobb et al.~\cite{cobb2013next} & Crop improvement & Defined next generation phenotyping requirements & Connected phenotyping to genotype-phenotype understanding & Data integration remained unresolved & Supports biological interpretation focus \\
2014 & Andrade-Sanchez et al.~\cite{andradesanchez2014phenotyping} & Field robotics & Field-based high throughput phenotyping platform & Enabled repeated crop measurements in field plots & Platform-specific workflow & Motivates mobile phenotyping infrastructure \\
2014 & Araus and Cairns~\cite{araus2014field} & Field HTPP & Positioned field phenotyping as breeding frontier & Linked sensing to selection under real environments & Translating data to knowledge remained hard & Supports breeding-facing PhenoAgent \\
2015 & Sankaran et al.~\cite{sankaran2015uav} & UAV sensing & UAV-based low-altitude phenotyping & Enabled rapid field trait acquisition & Limited under-canopy and root zone context & Shows strength and boundary of UAV phenotyping \\
2015 & Minervini et al.~\cite{minervini2015image} & Image analysis & Reviewed image-analysis approaches & Standardized image phenotyping concepts & Feature engineering and task specificity & Background for deep perception \\
2017 & Tardieu et al.~\cite{tardieu2017plant} & Sensors-to-knowledge & Emphasized conversion from sensors to knowledge & Shifted discourse beyond raw measurement & No operational agentic framework & Conceptual bridge to PhenoAgent \\
2017 & Pound et al.~\cite{pound2017deep} & Deep learning & CNN-based root and shoot feature localization & Demonstrated deep learning for trait extraction & Requires labeled, task-specific datasets & Foundation for automated organ perception \\
2017 & Ubbens and Stavness~\cite{ubbens2017deep} & Deep phenomics & Open deep learning platform & Made CNN phenotyping more accessible & Limited domain generalization & Supports reusable model pipelines \\
2017 & Coppens et al.~\cite{coppens2017unlocking} & Data integration & Highlighted integration of phenotyping datasets & Emphasized data-driven value extraction & Fragmented standards and metadata & Supports unified crop knowledge layer \\
2017 & Walter et al.~\cite{walter2017smart} & Smart farming & Linked smart farming with sustainability & Connected sensors, data, and agricultural decisions & Broad rather than phenotyping-specific & Frames impact of decision support \\
2017 & Wolfert et al.~\cite{wolfert2017big} & Big data & Reviewed smart farming data ecosystems & Highlighted big-data opportunities and governance & Limited biological model coupling & Supports data infrastructure layer \\
2017 & Virlet et al.~\cite{virlet2017fieldscanalyzer} & Field platform & Field Scanalyzer robotic platform & Automated repeated field measurements & High infrastructure cost & Demonstrates high-resolution field monitoring \\
2018 & Kamilaris and Prenafeta-Boldu~\cite{kamilaris2018deep} & Deep learning survey & Surveyed deep learning in agriculture & Mapped image, disease, yield, and detection tasks & Limited focus on agentic fusion & Establishes AI growth trajectory \\
2018 & Liakos et al.~\cite{liakos2018machine} & ML in agriculture & Reviewed machine learning applications & Connected ML with crop, water, soil, and livestock tasks & Heterogeneous validation quality & Provides broader ML context \\
2018 & Ferentinos~\cite{ferentinos2018deep} & Disease detection & CNN diagnosis of plant diseases & Demonstrated high classification performance & Controlled datasets and transfer limits & Shows need for deployment-aware models \\
2019 & Maes and Steppe~\cite{maes2019uav} & UAV remote sensing & Reviewed UAVs for precision agriculture & Summarized sensor payloads and plant traits & Data interpretation remains challenging & Supports platform-sensor mapping \\
2019 & Madec et al.~\cite{madec2019ear} & Organ detection & Deep learning for wheat ear density & Automated yield-component proxy & Organ-specific model & Supports trait-specific perception modules \\
2019 & Khaki and Wang~\cite{khaki2019yield} & Yield prediction & Deep neural networks for crop yield prediction & Showed promise of data-driven prediction & Requires robust environmental transfer & Supports predictive layer in PhenoAgent \\
2020 & Tausen et al.~\cite{tausen2020greenotyper} & Fixed cameras & Distributed low cost camera phenotyping & Enabled large image time series and segmentation & Mainly aboveground visual traits & Supports fixed greenhouse sensing \\
2020 & Perich et al.~\cite{perich2020canopy} & Thermal UAV & UAV canopy-temperature phenotyping & Addressed thermal image correction and stress traits & Thermal context remains difficult & Supports water-stress monitoring \\
2020 & David et al.~\cite{david2020global} & Benchmarking & Global Wheat Head Detection dataset & Advanced public object-detection benchmarks & Narrow crop/organ focus & Supports benchmark-based validation \\
2020 & Jiang and Li~\cite{jiang2020deep} & Review & Deep learning for plant phenotyping & Organized AI tasks and methods & Limited foundation model perspective & Provides review baseline \\
2020 & Papoutsoglou et al.~\cite{papoutsoglou2020miappe} & Metadata standards & MIAPPE 1.1 for phenomic datasets & Improved reusability and interoperability & Adoption and linkage to AI pipelines vary & Supports structured PhenoAgent memory \\
2021 & Xie and Li~\cite{xie2021uavmultisensor} & UAV platform & RGB, thermal, multispectral, LiDAR UAV system & Demonstrated integrated sensor acquisition & Requires downstream biological fusion & Supports multimodal data capture \\
2021 & Atefi et al.~\cite{atefi2021robotic} & Robotics review & Reviewed robotic HTPP technologies & Mapped robotic perception and mobility & Closed-loop action remains limited & Supports UGV/robotic PhenoAgent layer \\
2022 & Borhani et al.~\cite{borhani2022vit} & Vision transformers & ViT-based plant disease classification & Introduced transformer attention for plant disease & Dataset and speed limitations & Supports foundation model transition \\
2022 & Purcell and Neubauer~\cite{purcell2023digital} & Digital twins & State-of-the-art agriculture digital twin review & Defined DT opportunities and gaps & Many twins are digital shadows & Supports decision and simulation layer \\
2022 & Slob and Hurst~\cite{slob2022digital} & Greenhouse DT & Digital twins for agricultural greenhouses & Connected Industry 4.0 and greenhouse systems & Early-stage implementations & Supports greenhouse virtual layer \\
2023 & Cardellicchio et al.~\cite{cardellicchio2023yolo} & YOLO phenotyping & YOLOv5 tomato trait detection & Fast single-stage detection of traits & Trait-level, not system-level reasoning & Supports crop-specific detectors \\
2023 & Nguyen et al.~\cite{nguyen2023uavmultisensory} & Multimodal fusion & UAV hyperspectral, thermal, LiDAR, deep multitask learning & Predicted multiple maize traits from fused data & Limited seed/root zone/management integration & Key support for multimodal PhenoAgent \\
2023 & Parez et al.~\cite{parez2023greenvit} & Efficient ViT & GreenViT for disease detection & Demonstrated ViT efficiency in agriculture & Disease-focused task & Supports lightweight attention models \\
2023 & Li et al.~\cite{li2023pmvt} & Mobile ViT & PMVT lightweight disease diagnosis & Edge/mobile deployment orientation & Classification task only & Supports field-ready AI modules \\
2023 & Gole et al.~\cite{gole2023trincnet} & Lightweight transformer & TrIncNet for disease identification & Improved transformer efficiency & Limited biological explanation & Supports deployable perception \\
2023 & de Silva and Brown~\cite{desilva2023multispectral} & Multispectral ViT & ViT-CNN approaches for multispectral disease detection & Linked spectral bands and deep models & Narrow disease domain & Supports spectral foundation for multimodal AI \\
2023 & Yang et al.~\cite{yang2023leafseg} & Segmentation & YOLOv8 and improved DeepLabV3 leaf segmentation & Improved leaf-level extraction & Species and imaging dependency & Supports automated plant-part masks \\
2023 & Peladarinos et al.~\cite{peladarinos2023digital} & Digital twins & Comprehensive smart-agriculture DT review & Mapped IoT, simulation, and DT applications & Limited crop-specific validation & Supports DT decision dashboard framing \\
2023 & Cesco et al.~\cite{cesco2023smart} & Smart agriculture DT & Digital twins for sustainability & Connected DTs with nitrogen and management decisions & Full closed loop phenotyping still limited & Supports management-aware phenotyping \\
2023 & Kalyani et al.~\cite{kalyani2023cloudfog} & Cloud-fog-edge DT & DT deployment architecture & Addressed edge/cloud infrastructure & Biological models remain shallow & Supports scalable deployment layer \\
2024 & Murphy et al.~\cite{murphy2024deep} & Deep learning review & Reviewed deep learning in image based phenotyping & Synthesized current AI phenotyping progress & Agentic reasoning not yet central & Strong baseline for novelty comparison \\
2024 & Li et al.~\cite{li2024foundation} & Foundation models & Reviewed foundation models in smart agriculture & Identified opportunities for reusable AI & Agricultural grounding and trust remain open & Supports PhenoAgent foundation model layer \\
2024 & Meraj et al.~\cite{meraj2024survey} & Computer vision survey & Surveyed CV-based plant phenotyping & Broadly mapped segmentation and trait extraction & Integration beyond vision remains limited & Supports argument beyond CV-only reviews \\
2024 & Rahman et al.~\cite{rahman2024digitaltwin} & Greenhouse DT & Next-generation network and ML greenhouse twin & Combined networks, ML, and greenhouse management & Still implementation-specific & Supports real-time smart greenhouse management \\
2024 & Li and Zhang~\cite{li2024dcyolo} & Field detection & DC-YOLO for field plant detection & Lightweight real-time detection direction & Focused detection target & Supports robotic perception \\
2025 & Jin et al.~\cite{jin2025seed} & Seed phenotyping review & Deep learning for seed HTPP & Moves AI phenotyping to seed traits & Seed-to-plant integration still weak & Supports seed layer of PhenoAgent \\
2025 & Wang et al.~\cite{wang2025sensors} & Sensor trends & Image-based HTPP sensor and AI trends & Summarized current sensor-to-insight pipeline & Still not agentic or continuum-based & Supports transition to next generation section \\
2025 & Khuimphukhieo and Silva~\cite{khuimphukhieo2025uas} & UAS review & UAS-based field HTP as breeder toolbox & Summarized UAV use for breeding & Mostly aerial and plot level view & Supports field scalability discussion \\
2025 & Arshad et al.~\cite{arshad2025simulink} & Greenhouse DT & Simulink-driven smart greenhouse digital twin & Integrated CNN, IoT, and ML control & Prototype-level validation & Supports AI greenhouse control layer \\
2025 & DigiHortiRobot~\cite{digihortirobot2025} & Robotic DT & Hydroponic greenhouse robotic digital twin & Combines sensing, planning, and robotic automation & Conceptual/progressive implementation & Supports closed loop robotics direction \\
2026 & Zhang et al.~\cite{zhang2026ground} & UGV review & Ground mobile robots from perception-decision-action perspective & Frames robotics as closed loop phenotyping & Practical autonomy barriers remain & Direct bridge to PhenoAgent action layer \\
2026 & Guo et al.~\cite{zhou2026tomato} & Smart greenhouse & Digital framework for tomato cultivation & Combines IoT, AI, DT, and decision system & Crop- and facility-specific & Supports greenhouse productization \\
2026 & Ojo et al.~\cite{zhang2026cropdt} & Crop digital twin & Predictive growth monitoring and adaptive light control & Links edge AI, sensing, and control & CEA-specific validation & Supports adaptive intervention \\
2026 & Luo et al.~\cite{luo2026dynamic} & CEA digital twin & Dynamic adaptive modular DT for resource-efficient CEA & Focuses on resource-efficient controlled environments & Interoperability and broad deployment remain open & Supports dynamic PhenoAgent concept \\

\end{longtable}
\endgroup
\subsection{Digital Twins and Decision Dashboards}

Digital twins represent one of the most important current advances because they provide a conceptual bridge between physical crops, sensor data, predictive models, simulation, and decision support. In agriculture, digital twins have been proposed for field management, greenhouse climate control, irrigation, crop growth prediction, disease monitoring, machinery coordination, and resource optimization \cite{purcell2023digital,peladarinos2023digital,cesco2023smart}. In controlled environment agriculture, the digital twin concept is especially powerful because the physical system is partially controllable. Lighting, irrigation, fertigation, temperature, humidity, CO2, airflow, and crop spacing can be measured and adjusted with relatively short feedback cycles.

Decision dashboards are the visible interface through which digital phenotyping becomes useful to growers, breeders, and researchers. A dashboard can integrate sensor readings, images, predicted traits, alerts, treatment logs, growth curves, and yield forecasts. However, many dashboards are descriptive rather than explanatory. They show what happened but do not always explain why it happened, what evidence supports that explanation, how confident the model is, or what action should be tested next. Recent greenhouse digital twin studies using ML, IoT, cloud fog edge infrastructure, and adaptive control show that the field is moving toward prediction and intervention, but broad biological reasoning and cross crop generalization remain unresolved \cite{rahman2024digitaltwin,arshad2025simulink,zhang2026cropdt}.

For the PhenoAgent perspective, digital twins should be treated as more than dashboards or simulations. A useful crop twin should maintain a structured memory of plant identity, seed source, substrate condition, environmental exposure, management events, image derived traits, stress episodes, predictions, and interventions. The agentic layer can then ask biologically meaningful questions. For instance, determining whether reduced growth arises from seed vigor, root‑zone salinity, irrigation timing, heat stress, disease onset, or genotype sensitivity, whether additional data are required before intervention, and which management adjustment is most plausible under the current evidence. This moves the role of the digital twin from passive monitoring to active reasoning. Table \ref{tab:current_chronological_advances} highlights advances in high throughput, robotic, AI driven, multimodal, and digital twin phenotyping.

\subsection{Current Gaps: From Isolated Measurements to Integrated Biological Understanding}

Despite this progress, most current systems still operate as pipelines for measurement rather than reasoning. A camera system may estimate leaf area, a UAV workflow may calculate vegetation indices, a deep network may segment wheat heads, and a dashboard may display greenhouse temperature, but these outputs often remain separated by sensor type, crop stage, experiment, or decision context. The central challenge is therefore no longer only how to acquire more data, but how to convert multimodal and temporal data into biologically meaningful, actionable crop intelligence. The current literature shows remarkable progress in sensors, platforms, AI models, and dashboards, but it also reveal limitations. For instance, a UAV study may be optimized for canopy indices, a robot may be optimized for navigation and image capture, a deep network may be optimized for segmentation accuracy, and a digital twin may be optimized for climate simulation. However these achievements are valuable, yet they do not fully solve the biological question of how seed, soil or substrate, plant development, environment, and management jointly produce yield, quality, and stress resilience. This gap has been recognized indirectly in work on phenotyping data integration, MIAPPE metadata, big data smart farming, digital twin interoperability, and AI review literature \cite{coppens2017unlocking,papoutsoglou2020miappe,wolfert2017big,meraj2024survey}.

In addition, current systems also struggle with generalization. Models trained on one crop, sensor, location, greenhouse, genotype set, imaging angle, or growth stage may not transfer to another. Foundation models and promptable segmentation can reduce this problem, but agriculture requires domain-specific grounding in plant morphology, physiology, phenology, environment, and management \cite{kirillov2023segment,li2024foundation,murphy2024deep}. Current models also need better uncertainty estimation, human in the loop correction, standardized metadata, robust benchmarking, and mechanisms for asking for additional measurements when evidence is insufficient. Without these capabilities, AI phenotyping can produce accurate looking outputs that are difficult to trust for breeding or production decisions. 

The most important unresolved gap is the absence of an integrated biological reasoning layer. High throughput phenotyping should not only say that a plant is smaller, hotter, more yellow, or lower yielding. It should explain whether the likely cause is seed vigor, substrate EC, poor germination, moisture deficit, nutrient imbalance, heat stress, disease progression, genotype sensitivity, or management timing. It should connect early seed and germination traits with later crop outcomes, link aboveground symptoms with belowground and environmental context, and recommend the next measurement or intervention. This is the conceptual space for PhenoAgent, a multimodal, agentic phenotyping framework that builds on current sensors, robotics, AI models, and digital twins, but organizes them around crop understanding and decision making. 

\begin{table}[h]
\centering
\caption{Existing limitations in AI-driven phenotyping and how they inform the need for a PhenoAgent-based framework. The table translates the literature review into design requirements for the next generation of phenotyping.}
\label{tab:current_gaps_phenoagent_requirements}
\scriptsize
\setlength{\tabcolsep}{2.2pt}
\renewcommand{\arraystretch}{1.17}
\resizebox{\textwidth}{!}{%
\begin{tabular}{P{3.0cm} P{3.5cm} P{3.5cm} P{3.4cm} P{3.4cm}}
\toprule
\textbf{Current Capability} & \textbf{What current systems do well} & \textbf{Main unresolved gap} & \textbf{Design requirement for PhenoAgent} & \textbf{Supporting literature} \\
\midrule
High throughput sensing & Rapid imaging, repeated measurements, spectral and structural traits & Sensor outputs remain fragmented by modality and platform & Time-synchronized sensor fusion by plant, plot, treatment, and growth stage & Tardieu et al.~\cite{tardieu2017plant}; Wang et al.~\cite{wang2025sensors} \\
Robotic and UAV platforms & Scalable field or greenhouse data acquisition & Limited biological interpretation and closed loop action & Platform-aware sensing plans and autonomous follow up measurements & Atefi et al.~\cite{atefi2021robotic}; Zhang et al.~\cite{zhang2026ground}; Khuimphukhieo and Silva~\cite{khuimphukhieo2025uas} \\
Deep learning perception & Detection, segmentation, classification, and regression & Task-specific models with limited transfer and weak causal explanation & Foundation-model perception with uncertainty and human correction & Pound et al.~\cite{pound2017deep}; Murphy et al.~\cite{murphy2024deep}; Li et al.~\cite{li2024foundation} \\
Multimodal fusion & Improved prediction by combining spectral, structural, thermal, and environmental data & Fusion often remains feature-level rather than biology-level & seed soil plant environment management representation learning & Nguyen et al.~\cite{nguyen2023uavmultisensory}; Coppens et al.~\cite{coppens2017unlocking}; Papoutsoglou et al.~\cite{papoutsoglou2020miappe} \\
Digital twins and dashboards & Monitoring, simulation, prediction, and user visualization & Many implementations are descriptive dashboards or digital shadows & Agentic digital twin with reasoning, evidence tracking, and decision support & Purcell and Neubauer~\cite{purcell2023digital}; Rahman et al.~\cite{rahman2024digitaltwin}; Ojo et al.~\cite{zhang2026cropdt} \\
Decision support & Alerts, trends, forecasts, and rule-based recommendations & Weak explanation of causes and next-best measurements & Explainable recommendations grounded in crop physiology and trial context & Walter et al.~\cite{walter2017smart}; Cesco et al.~\cite{cesco2023smart}; Arshad et al.~\cite{arshad2025simulink} \\
\bottomrule
\end{tabular}}
\end{table}

Current high throughput and AI driven phenotyping has substantially reduced many of the data acquisition and perception barriers that limited conventional phenotyping. The field now has sensors, robots, UAVs, image models, foundation model concepts, fusion methods, and digital dashboards. The remaining challenge is to connect these components into a coherent biological reasoning system that can explain crop response and support decisions across the seed soil plant continuum. This gap provides the direct transition to the next section, where PhenoAgent is introduced as a next generation framework for multimodal crop intelligence. Table \ref{tab:current_gaps_phenoagent_requirements} highlights in AI-driven phenotyping and how they inform the need for a PhenoAgent-based framework. 

\section{Next-Generation Phenotyping}
\label{sec-4}
The preceding sections show that plant phenotyping has progressed from manual scoring to high-throughput sensing, robotic data acquisition, deep learning, multimodal fusion, and digital visualization. However, next-generation phenotyping requires more than faster measurement or dashboard reporting. It requires reasoning frameworks that can interpret crop status, connect observations over time, infer plausible causes, and support management decisions. 
\begin{figure}[h!]
    \centering
    \includegraphics[width=1\textwidth]{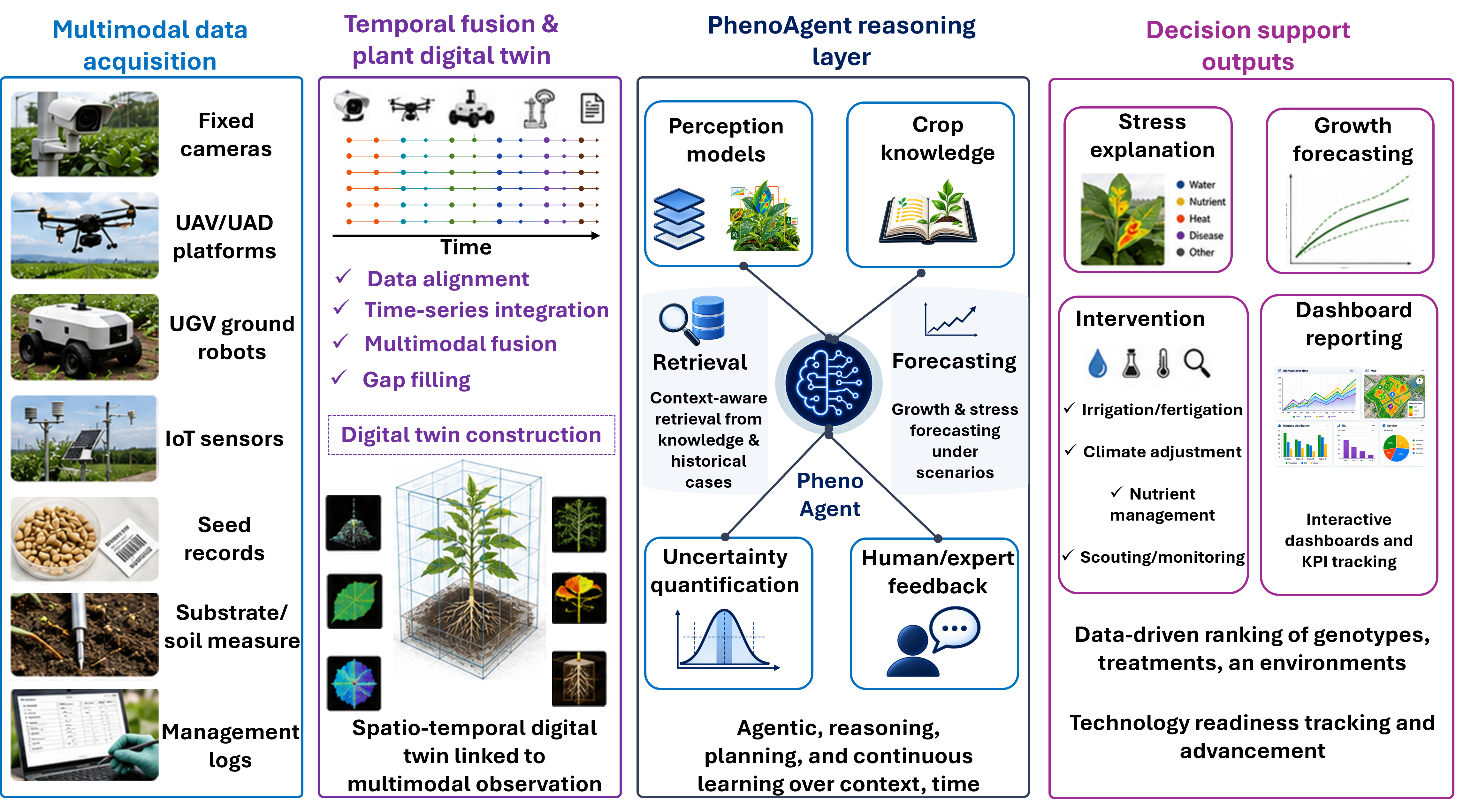}
    \caption{Core components of PhenoAgent as a next-generation phenotyping intelligence layer that links multimodal data capture, temporal fusion, plant digital twins, agentic reasoning, and decision-ready outputs. It integrates visual, environmental, substrate, seed, and management data to explain stress, predict growth, recommend interventions, support dashboards, rank trials, and guide technology translation.}
    \label{fig:phenoagent_graphical_abstract}
\end{figure}

\subsection{PhenoAgent Conceptual Framework}

In this review, \textit{PhenoAgent} is introduced as a conceptual framework for agentic multimodal crop intelligence. It seeks to advance phenotyping from standalone measurements to a reasoning layer that can relate crop observations to biological context, management history, and decision support. Figure~\ref{fig:phenoagent_graphical_abstract} summarizes the core elements of this framework. PhenoAgent integrates seed information, soil/substrate and root-zone measurements, plant traits, environmental conditions, and management actions into a time-resolved crop-state representation. This representation enables crop development to be understood as a progression of evidence rather than as disconnected sensor outputs or isolated trait values. The rationale behind PhenoAgent is that crop performance is rarely explained by a single measurement. For example, reduced growth can be associated with low seed vigor, delayed emergence, water or nutrient imbalance in the root zone, salinity, disease pressure, heat stress, genotype-specific sensitivity, or the accumulated effect of earlier management actions. Current phenotyping systems can already segment organs, estimate canopy traits, compute vegetation indices, detect stress symptoms, and visualize sensor trends. However, many of these systems still stop at measurement or monitoring. Practical crop decisions require a more integrated interpretation: what is happening, why it may be happening, how confident the system is, what evidence is missing, and what action should be considered next. PhenoAgent therefore extends the phenomics objective of moving from sensors to knowledge towards evidence-based reasoning, human-in-the-loop verification, and closed-loop decision support \cite{tardieu2017plant,pieruschka2019plant,murphy2024deep}.

The conceptual framework shown in Figure~\ref{fig:phenoagent_graphical_abstract} is structured into four interlinked layers. The first layer focuses on multimodal data collection: fixed cameras, UAV/UAD systems, UGVs, IoT sensors, seed records, substrate or root-zone measurements, and management logs each contribute complementary evidence about the crop system. The second layer addresses temporal fusion and the creation of a plant digital twin, in which diverse data streams are synchronized over time and associated with plant, tray, plot, or greenhouse-zone identifiers. The third layer is the PhenoAgent reasoning component, integrating perception models, agronomic knowledge, retrieval of prior evidence, forecasting, uncertainty quantification, and human/expert feedback. The fourth layer provides decision support, converting these interpretations into stress diagnoses, growth predictions, intervention suggestions, dashboard outputs, genotype or treatment prioritization, and tracking of technology readiness.

\begin{figure}[h!]
    \centering
    \includegraphics[width=0.7\textwidth]{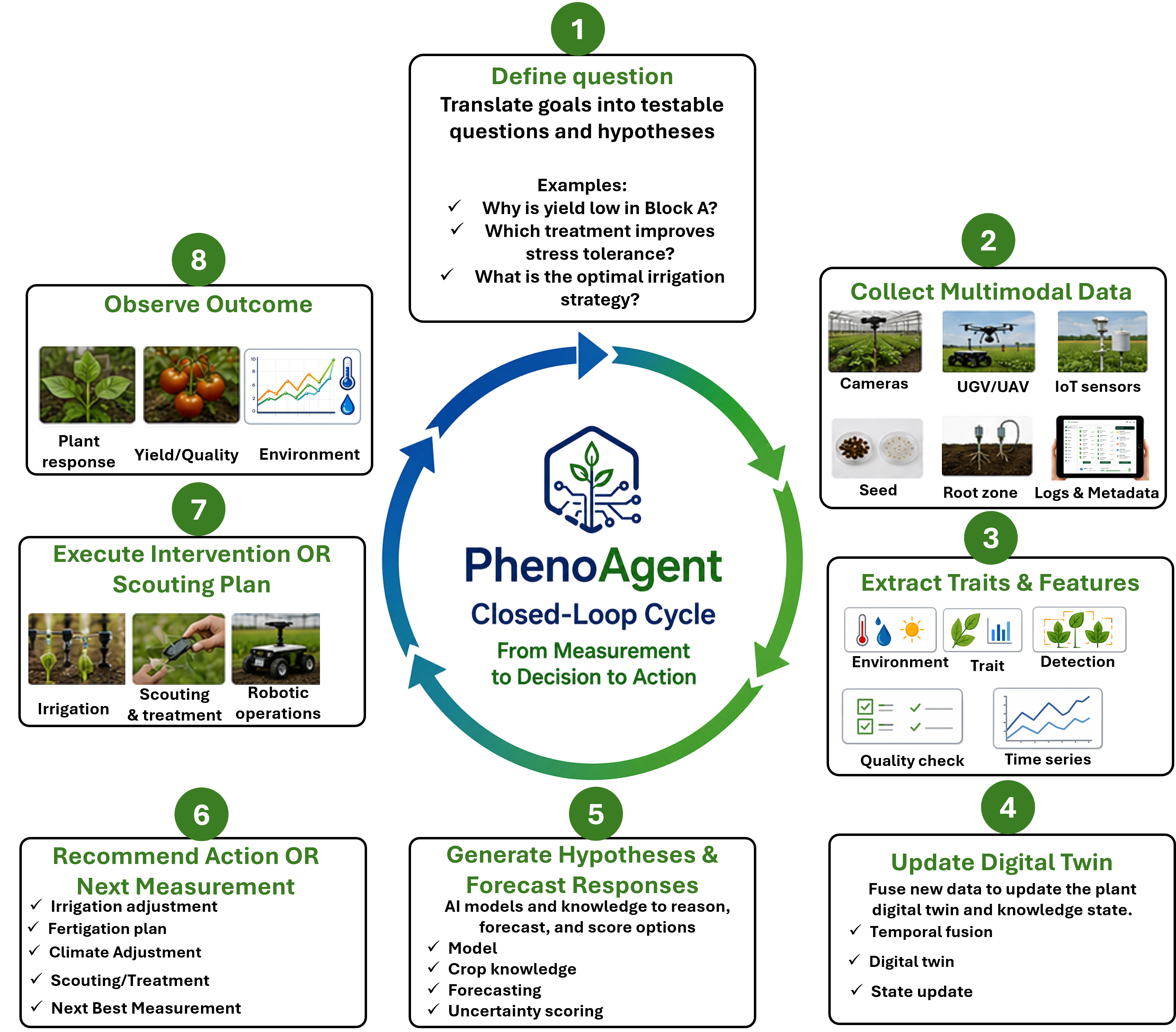}
    \caption{Closed-loop PhenoAgent workflow from phenotyping question to crop decision and outcome evaluation.}
    \label{fig:phenoagent_closed_loop_cycle}
\end{figure}

\begin{table*}[t!]
\centering
\caption{Algorithmic workflow of PhenoAgent from multimodal crop observation to decision support.}
\label{tab:phenoagent_conceptual_workflow}
\scriptsize
\setlength{\tabcolsep}{2.4pt}
\renewcommand{\arraystretch}{1.18}
\resizebox{\textwidth}{!}{%
\begin{tabular}{M{0.7cm} P{3.0cm} P{4.0cm} P{4.2cm} P{3.8cm}}
\toprule
\textbf{Step} & \textbf{Stage} & \textbf{Main inputs} & \textbf{Core function} & \textbf{Expected output} \\
\midrule
1 & Phenotyping context definition & Crop type, growth stage, target trait, treatment, management objective, and trial metadata & Define what should be monitored, why it matters, and how crop units will be tracked over time & Crop-unit identity, trial context, and phenotyping objective \\

2 & Multimodal data acquisition & Fixed cameras, UAV/UAD platforms, UGV robots, IoT sensors, seed records, substrate/root-zone measurements, and management logs & Collect complementary visual, environmental, biological, and management evidence from the crop system & Raw multimodal crop evidence \\

3 & Temporal alignment and quality control & Images, sensor logs, timestamps, crop-unit IDs, and metadata & Synchronize data streams, remove low-quality records, align observations with plant/tray/plot identity, and organize measurements along the crop timeline & Clean time-aligned evidence record \\

4 & Trait and state extraction & RGB/depth/thermal/spectral images, IoT signals, substrate data, and management events & Estimate crop traits and state indicators such as emergence, vigor, canopy growth, stress symptoms, disease cues, root-zone condition, and yield-related proxies & Trait/state profile with confidence values \\

5 & Plant digital twin construction & Trait profile, environmental history, root-zone status, seed information, and management records & Build a time-resolved crop-state representation that links phenotype, environment, substrate/root-zone condition, and management history & Dynamic plant/tray/plot digital twin \\

6 & PhenoAgent reasoning & Digital twin, perception outputs, crop knowledge, retrieved evidence, forecasting models, uncertainty estimates, and expert feedback & Interpret the crop state, identify likely stress drivers, compare alternative explanations, and determine whether additional evidence is needed & Explainable crop-state interpretation \\

7 & Forecasting and decision support & Current crop state, historical trends, candidate actions, uncertainty, and grower/breeder constraints & Predict likely crop trajectory and rank suitable actions or additional measurements under the current conditions & Growth/stress forecast and recommended next steps \\

8 & Human review and closed-loop update & Expert assessment, manual checks, intervention outcomes, and new observations & Validate or revise the recommendation, record outcomes, and update the crop-state memory for future reasoning & Audited decision record and improved system knowledge \\
\bottomrule
\end{tabular}}
\end{table*}

\subsection{PhenoAgent Conceptual Workflow}
Figure~\ref{fig:phenoagent_closed_loop_cycle} further explains how PhenoAgent can operate as a closed-loop phenotyping system. This second figure presents a practical use case in which the process starts with a testable agronomic question, such as why yield is low in a specific block, which treatment improves stress tolerance, or what irrigation strategy is most suitable under current conditions. The system then collects multimodal evidence, extracts traits and time-series features, updates the plant digital twin, and uses the reasoning layer to generate hypotheses and forecast likely crop responses. Based on this interpretation, PhenoAgent can recommend either a management action, such as irrigation, fertigation, climate adjustment, or scouting, or a next measurement when the available evidence is insufficient. After the action or measurement is completed, the observed crop response is used to update the digital twin and improve the next reasoning cycle.
This closed-loop structure is important because phenotyping becomes more useful when it is connected to action and feedback. A one-time prediction or stress alert may support monitoring, but it does not by itself create an adaptive crop-management system. In contrast, a PhenoAgent-style workflow treats each recommendation as part of a learning cycle: the system observes the crop, interprets the evidence, suggests an action or additional measurement, evaluates the outcome, and updates its crop-state memory. Table~\ref{tab:phenoagent_conceptual_workflow} further summarizes this workflow in algorithmic form, showing how heterogeneous crop data can be transformed into aligned evidence, trait profiles, digital-twin updates, explainable hypotheses, forecasts, recommendations, and outcome-based learning within a closed-loop phenotyping framework.

\begin{table*}[t!]
\centering
\caption{Conceptual distinction between current AI phenotyping systems and the suggested PhenoAgent framework. The table highlights why PhenoAgent should be treated as a next generation direction rather than another image-analysis pipeline.}
\label{tab:phenoagent_conceptual_distinction}
\scriptsize
\setlength{\tabcolsep}{2.2pt}
\renewcommand{\arraystretch}{1.15}
\resizebox{\textwidth}{!}{%
\begin{tabular}{P{3.0cm} P{3.8cm} P{4.0cm} P{4.2cm} P{3.0cm}}
\toprule
\textbf{Dimension} & \textbf{Current Common Practice} & \textbf{PhenoAgent Direction} & \textbf{Expected Advantage} & \textbf{Key References} \\
\midrule
System goal & Estimate traits, indices, or labels from images and sensors & Explain crop state and recommend evidence-aware next steps & Moves from measurement to actionable intelligence & Tardieu et al.~\cite{tardieu2017plant}; Murphy et al.~\cite{murphy2024deep} \\
Data organization & Separate image folders, sensor logs, trial sheets, and dashboard values & Unified seed soil plant environment management time series & Enables causal interpretation across crop development & Coppens et al.~\cite{coppens2017unlocking}; Papoutsoglou et al.~\cite{papoutsoglou2020miappe} \\
AI model role & Task-specific detection, segmentation, classification, or regression & Modular perception plus foundation model and retrieval-assisted reasoning & Supports reusable and context-aware interpretation & Bommasani et al.~\cite{bommasani2021foundation}; Li et al.~\cite{li2024foundation} \\
Decision support & Threshold alerts, plots, and descriptive dashboards & Hypothesis ranking, uncertainty, intervention recommendation, and follow up sensing & Improves grower and breeder decision confidence & Walter et al.~\cite{walter2017smart}; Rahman et al.~\cite{rahman2024digitaltwin} \\
Human role & Manual labeling, scouting, interpretation, and final decision & Expert guided validation, feedback, correction, and approval of interventions & Preserves agronomic expertise while reducing routine burden & Wolfert et al.~\cite{wolfert2017big}; Peladarinos et al.~\cite{peladarinos2023digital} \\
Translation path & Research prototype or sensor-specific system & Scalable proof of concept progressing toward high-TRL deployment & Converts data infrastructure into deployable phenotyping product & Atefi et al.~\cite{atefi2021robotic}; DigiHortiRobot~\cite{digihortirobot2025} \\
\bottomrule
\end{tabular}}
\end{table*}

\subsection{Conventional Phenotyping vs. PhenoAgent}

A key limitation of conventional phenotyping and many high-throughput systems is that measurement alone does not ensure understanding. Manual scoring, RGB, thermal, hyperspectral, depth, and IoT monitoring each capture only part of the crop system: RGB shows size, color, and visible symptoms; thermal indicates canopy temperature and water stress; hyperspectral detects subtle biochemical or physiological changes; and IoT sensors track root-zone and greenhouse conditions. Because each modality is partial, crop and breeding decisions require integrating these observations over time and within biological and management context. The PhenoAgent idea differs from conventional, measurement-centered phenotyping by treating phenotyping as reasoning rather than collecting isolated trait values. Beyond reporting that a plant is smaller, warmer, slower growing, or visually stressed, it evaluates plausible causes such as poor seed vigor, delayed emergence, nutrient imbalance, water limitation, salinity, disease or pests, heat stress, genotype-environment interaction, or mistimed management \cite{furbank2011phenomics,coppens2017unlocking,wolfert2017big}. Phenotyping thus moves from observation to inference, linking measurements to likely explanations and next checks or actions.

This transition aligns with broader AI trends. Deep learning improved vision by learning task-relevant image features, while transformers and foundation models advanced transferable representations and prompt-based interaction \cite{dosovitskiy2021vit,bommasani2021foundation,li2024foundation}. More recently, retrieval-augmented generation and agentic tool use have shown how models can be combined with external knowledge, structured data, reasoning, and task-specific tools \cite{lewis2020rag,yao2023react,schick2023toolformer}. In agriculture, early retrieval-based advisors and digital twins are starting to link literature, sensors, simulations, and recommendations, but remain weakly integrated with plant-level multimodal phenotyping and robotic data collection \cite{vizniuk2025ragagri,sawant2026rag,han2026goat}. Table~\ref{tab:phenoagent_conceptual_distinction} summarizes the distinction between conventional or current AI-enabled phenotyping and the PhenoAgent framework. The comparison highlights that PhenoAgent is not simply another image-analysis pipeline. It represents a shift toward temporally grounded, multimodal, explainable, and decision-oriented crop intelligence.


\section{Research Gaps, Deployment Barriers, and Future Opportunities}
\label{sec-5}

\begin{figure}[h!]
    \centering
    \includegraphics[width=1\textwidth]{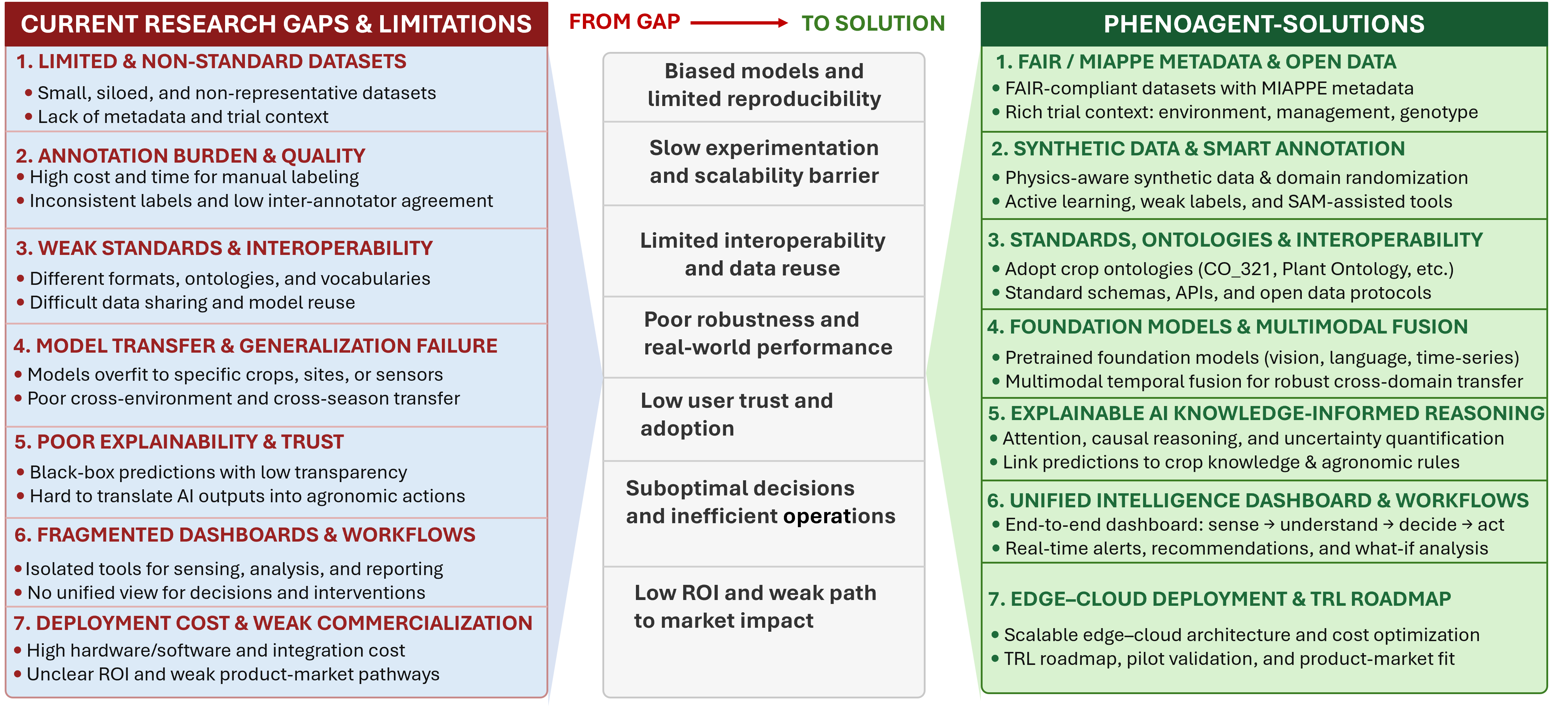}
    \caption{Research gap-to-solution map for next generation automated phenotyping.}
    \label{fig:discussion_gap_to_solution_map}
\end{figure}

Advances in high-throughput sensing, robotics, deep learning, foundation models, digital twins, and dashboards have enabled automated plant phenotyping, but the field is at a transition point. Many platforms can capture large visual, spectral, thermal, structural, and environmental datasets, yet few convert them into robust, explainable, biologically valid, and deployable decisions. The focus must shift beyond accuracy and sensor specs to practical next-generation requirements \cite{tardieu2017plant,papoutsoglou2020miappe,murphy2024deep}. A PhenoAgent-style ecosystem can unify fragmented progress into a product pathway by building multimodal reasoning that links seed, soil/substrate, plant, environment, and management. The translational goal is to turn research phenotyping into decision-support for breeding, greenhouse production, seed testing, substrate optimization, precision irrigation, and autonomous scouting. This section summarizes key limitations/gaps and opportunities to move phenotyping from prototypes to high-TRL agricultural products as outlines in Figure~\ref{fig:discussion_gap_to_solution_map}.

\subsection{Current Limitations: Datasets, Annotation, Benchmarking, and Standardization}
Dataset scarcity is a major barrier to AI-driven phenotyping. Plant phenotyping datasets are hard to build because crop appearance varies with genotype, growth stage, environment, viewpoint, illumination, sensor, background, treatment, and management. They also require biological metadata such as crop identity, genotype/cultivar, seed lot, substrate/soil, irrigation/fertigation history, environmental exposure, growth stage, treatment protocol, sampling frequency, and ground-truth method—without which models may fit images but fail at biological interpretation and reproducibility \cite{coppens2017unlocking,papoutsoglou2020miappe,jin2025seedphenotyping}.
Annotation is another bottleneck: segmentation masks, organ labels, disease scores, counts, root traits, stress grades, and yield components demand expert effort and can differ across annotators. Benchmarks like Global Wheat Head Detection help standardize evaluation for specific traits, but reviews still call for larger, more diverse, better-annotated datasets \cite{david2020global,jiang2020deep,murphy2024deep}. Promptable segmentation, weak supervision, active learning, synthetic data, and sim-to-real can reduce labor, but still require biologically meaningful validation. Synthetic data help only if they capture realistic geometry, sensor artifacts, growth stages, and environmental variation \cite{kirillov2023segment,liu2025mars}.
Data standardization connects dataset creation to reusable intelligence. MIAPPE guides phenotyping metadata, yet many pipelines still separate images, sensor logs, management records, and labels \cite{papoutsoglou2020miappe}. Future PhenoAgent systems should treat dataset design as a scientific contribution: machine-readable, temporally aligned data linked to biological identity and auditable from raw measurement to model output. This enables cross-crop/environment comparisons, reproducibility, digital twins, and multi-institution benchmarking. Table \ref{tab:discussion_research_gaps} summarizes key gaps and opportunities.

\begin{table*}[h!]
\centering
\caption{Key research gaps in automated phenotyping and corresponding future opportunities for PhenoAgent-style systems.}
\label{tab:discussion_research_gaps}
\scriptsize
\setlength{\tabcolsep}{2.2pt}
\renewcommand{\arraystretch}{1.16}
\resizebox{\textwidth}{!}{%
\begin{tabular}{P{3.1cm} P{4.0cm} P{4.0cm} P{3.9cm} P{2.8cm}}
\toprule
\textbf{Gap} & \textbf{Current challenge} & \textbf{Future opportunity} & \textbf{Validation requirement} & \textbf{Key references} \\
\midrule
Dataset scarcity & Few datasets cover multiple crops, sensors, stages, environments, and management histories & Longitudinal seed soil plant datasets with real and synthetic data & Cross-site benchmark splits and metadata completeness checks & Papoutsoglou et al.~\cite{papoutsoglou2020miappe}; Liu et al.~\cite{liu2025mars} \\
Annotation burden & Expert labels are expensive and inconsistent across traits and crops & Promptable segmentation, active learning, weak labels, expert-in-the-loop correction & Label agreement, uncertainty reporting, and audit trails & Kirillov et al.~\cite{kirillov2023segment}; Murphy et al.~\cite{murphy2024deep} \\
Benchmarking & Many studies report isolated accuracy on private datasets & Open protocols for detection, trait extraction, stress prediction, and decision value & Standard metrics for accuracy, robustness, calibration, and agronomic utility & David et al.~\cite{david2020global}; Jiang and Li~\cite{jiang2020deep} \\
Data standards & Image, sensor, environment, and management records are often disconnected & FAIR and MIAPPE-aligned multimodal phenotyping records & Reusable schemas linked to plant/tray/plot identity and timestamps & Coppens et al.~\cite{coppens2017unlocking}; Papoutsoglou et al.~\cite{papoutsoglou2020miappe} \\
Biological reasoning & Models detect patterns but often do not explain crop causes & Agentic multimodal reasoning with crop knowledge and digital twins & Expert agreement, causal plausibility, and intervention outcomes & Tardieu et al.~\cite{tardieu2017plant}; Li et al.~\cite{li2024foundation} \\
\bottomrule
\end{tabular}}
\end{table*}

\subsection{Current Limitations: Generalization Across Crops and Environments}
Generalization is the difference between a promising research model and a deployable phenotyping system. A model trained on one crop, greenhouse, camera, season, disease dataset, or growth stage may perform poorly when lighting changes, plants overlap, genotypes differ, sensors drift, or management regimes shift. This problem is especially strong in agriculture because phenotype is not static. A seedling, vegetative plant, flowering plant, and fruiting plant can look like different visual domains even when they belong to the same crop. Similarly, the same stress can appear differently depending on genotype, substrate, irrigation history, and climate conditions \cite{araus2014field,poorter2016pampered,murphy2024deep,hossen2025transferlearningagriculture}.
Foundation models offer a partial solution by learning reusable representations across large datasets, while vision transformers and promptable models can improve flexibility in segmentation and recognition tasks \cite{dosovitskiy2021vit,bommasani2021foundation,li2024foundation}. However, agricultural foundation models must be grounded in plant specific variation. General purpose models may segment visible objects but may not understand growth stage, leaf hierarchy, crop architecture, disease progression, root zone stress, or management context. Future work should therefore develop plant foundation models trained on diverse visual, spectral, temporal, and environmental data, combined with domain adaptation methods that can adjust to new crops, sensors, facilities, and field conditions.
For PhenoAgent, generalization should be evaluated at three levels. The first is perception generalization for instance whether segmentation, detection, and trait extraction remain accurate across sensors and crops. The second is temporal generalization in which  growth forecasting and anomaly detection remain valid across seasons, seed lots, and management regimes. The third is decision generalization in which recommended actions remain biologically plausible and operationally useful under different greenhouse, field trail, and breeding contexts.

\subsection{Current Limitations: Biological Validity, Explainability, and Trust}

Biological validity is the core requirement that distinguishes plant phenotyping from generic image analysis. A model output is not valuable only because it is numerically accurate; it must correspond to a meaningful biological trait, stress process, growth stage, or management decision. For example, a canopy cover estimate is useful only if it reflects true growth differences rather than camera angle, lighting, or segmentation bias. A disease label is useful only if it is linked to real symptoms, severity, uncertainty, and appropriate follow up. A yield forecast is useful only if it is interpretable enough for breeders or growers to act on it \cite{tardieu2017plant,pieruschka2019plant,watt2020phenotyping}. Explainability is therefore not a cosmetic feature. It is a requirement for trust, debugging, scientific interpretation, and deployment. In PhenoAgent style systems, explanations should include the evidence used, the missing evidence, the alternative hypotheses considered, the confidence level, and the recommended next measurement or intervention. This is particularly important for agentic systems and LLM supported decision tools, where hallucination, overconfidence, and weak grounding can create risk \cite{bommasani2021foundation,lewis2020rag,vizniuk2025ragagri}. Retrieval augmented and tool using agents can reduce some of this risk by grounding outputs in records, sensor data, trial histories, and agronomic references, but the reasoning must remain auditable and human supervised. Trust also depends on calibration and failure awareness. A phenotyping system should know when it is outside its training distribution, when sensor quality is poor, when labels are uncertain, and when expert inspection is needed. Human in the loop design should be integrated into the system so agronomists and breeders can correct model outputs, validate alerts, and approve recommendations. This is the route by which AI phenotyping can become a trusted collaborator rather than a black box scorer.

\subsection{Current Limitations: Edge Cloud Deployment and Data Governance}

Deployment is a major barrier because automated phenotyping systems must operate under real agricultural constraints including heat, humidity, dust, variable light, sensor drift, unstable connectivity, limited power, greenhouse workflow interruptions, crop cycle deadlines, and maintenance constraints. Edge AI can support low latency inference for segmentation, detection, anomaly alerts, and robot navigation, while cloud computing can support heavy model training, historical data analysis, dashboard services, and multi site learning. Cloud fog edge digital twin literature shows that scalable smart agriculture systems require careful architecture choices, not only accurate models \cite{kalyani2023cloudfog,rahman2024digitaltwin,upadhyay2025plantdiseasecv}.

Data ownership and governance are equally important. Phenotyping data may include commercially sensitive information about seed lines, greenhouse productivity, irrigation strategy, yield performance, disease incidence, and management practices. Growers and breeding companies may hesitate to share data unless ownership, access control, privacy, cybersecurity, and model use rights are clearly defined. smart farming big data reviews have identified governance and value sharing as central challenges in agricultural data ecosystems \cite{wolfert2017big,walter2017smart,duguma2025iotagriculture,kizielewicz2025uavsensors}. A deployable PhenoAgent system should therefore include clear policies for data access, local storage, cloud synchronization, anonymization, audit logs, and exportable trial records.

Scalable deployment also requires maintainability. Sensor calibration, camera cleaning, robot charging, data backups, model updates, annotation review, and dashboard uptime must be treated as operational requirements. A system that performs well in a laboratory but requires constant expert engineering support may not reach high TRL. Future platforms should therefore use modular hardware, standardized APIs, robust data pipelines, automated quality checks, and user-centered dashboards that fit existing grower and breeder workflows. Table \ref{tab:discussion_deployment_barriers} highlights deployment barriers and practical design responses for scalable automated phenotyping systems. 

\begin{table*}[h!]
\centering
\caption{Deployment barriers and practical design responses for scalable automated phenotyping systems.}
\label{tab:discussion_deployment_barriers}
\scriptsize
\setlength{\tabcolsep}{2.1pt}
\renewcommand{\arraystretch}{1.16}
\resizebox{\textwidth}{!}{%
\begin{tabular}{P{3.0cm} P{3.8cm} P{4.0cm} P{3.9cm} P{3.0cm}}
\toprule
\textbf{Barrier} & \textbf{Why it matters} & \textbf{Design response} & \textbf{TRL validation metric} & \textbf{Related references} \\
\midrule
Sensor robustness & Dust, humidity, lighting, drift, and occlusion reduce data quality & Calibration routines, QA flags, redundant sensing, automated health checks & Uptime, missing-data rate, calibration error & Xie and Li~\cite{xie2021uavmultisensor}; Nguyen et al.~\cite{nguyen2023uavmultisensory} \\
Compute constraints & Real-time alerts and robotics need low latency, while training requires scale & Edge inference for alerts; cloud for training, dashboard, and historical analysis & Latency, throughput, energy use, cloud cost & Kalyani et al.~\cite{kalyani2023cloudfog}; Rahman et al.~\cite{rahman2024digitaltwin} \\
Model maintenance & Crop cycles, sensors, and facilities change over time & Continuous validation, drift monitoring, versioned models, human correction & Degradation rate and retraining interval & Murphy et al.~\cite{murphy2024deep}; Li et al.~\cite{li2024foundation} \\
Data governance & Trial and yield records may be commercially sensitive & Access control, local/cloud policy, anonymization, audit logs, ownership agreement & Compliance, user acceptance, exportability & Wolfert et al.~\cite{wolfert2017big}; Walter et al.~\cite{walter2017smart} \\
Workflow integration & Systems fail if they interrupt grower or breeder routines & Dashboard co-design, role-specific views, mobile alerts, simple feedback tools & Usability, adoption, decision response time & Peladarinos et al.~\cite{peladarinos2023digital}; Cesco et al.~\cite{cesco2023smart} \\
Commercial readiness & Research prototypes often lack support, reliability, and business model & TRL staging, pilot validation, service model, training, maintenance plan & TRL milestone completion and decision-value evidence & DigiHortiRobot~\cite{digihortirobot2025}; Zhang et al.~\cite{zhang2026ground} \\
\bottomrule
\end{tabular}}
\end{table*}

\subsection{Future Opportunity: High TRL Translation for Commercialization}

Automated phenotyping systems must reduce labor, improve trial performance, detect stress earlier, improve crop uniformity, support seed and substrate selection, increase yield predictability, reduce resource waste, or shorten breeding cycles. The commercial value of automated phenotyping lies less in the sensing hardware itself and more in the reliability of the decisions that the system enables. This distinction is important because many phenotyping systems are technically impressive but too expensive, too complex, or too narrow for routine deployment \cite{motta2025coffeeclassification}.
The market context supports high TRL development. Public market estimates vary by methodology, but consistently show that plant phenotyping is a growing specialized segment connected to much larger markets in smart greenhouses, precision agriculture, and agricultural robotics. For example, recent market reports estimate plant phenotyping in the hundreds of millions of dollars by 2030, while smart greenhouse markets are commonly projected in the multi billion dollar range by 2030 and agricultural robotics forecasts are substantially larger \cite{mordor2025phenotyping,marketresearchreports2024phenotyping,gia2025smartgreenhouse,grandview2025agrobots}. These figures should not be treated as exact scientific measurements but are useful for positioning automated phenotyping as a strategic enabling technology for the commercialization of AgriTech.

For a PhenoAgent type product, the commercialization pathway can be organized into three product layers. The first layer is the data infrastructure that includes sensors, robots, cameras, IoT devices, and integration services. The second layer is intelligence software consisting of data synchronization, trait extraction, digital twin records, forecasting, explainability, and recommendation engines. The third layer is the decision service that includes dashboards, trial analytics, grower alerts, breeding reports, and subscription based monitoring. The strongest product opportunity is likely to emerge from the software and decision service layers because hardware costs may decline over time, while value increasingly comes from proprietary datasets, validated models, workflow integration, and trusted recommendations. Figure \ref{fig:current_high_throughput_phenotyping_ecosystem} summarizes the high TRL translation pathway for PhenoAgent style crop intelligence.

\subsection{Roadmap: Agentic Crop Intelligence}
The future roadmap for automated phenotyping can be organized around four converging technologies. The first is plant foundation models. These models should learn from diverse images, spectra, point clouds, time series sensor data, management logs, textual protocols, and experimental metadata. Their goal should not be generic visual recognition alone, but crop aware representation learning that supports segmentation, trait extraction, growth stage recognition, stress diagnosis, and transfer across crops and facilities \cite{bommasani2021foundation,li2024foundation,murphy2024deep,islam2025smartdigitaltwins}. The second technology is the plant digital twin. Future digital twins should represent crop units as dynamic biological states, not only as environmental simulations or dashboards. A useful digital twin should link seed and genotype information, substrate/root zone conditions, visual phenotype, environmental exposure, management history, predicted growth, stress risk, and intervention outcomes \cite{purcell2023digital,rahman2024digitaltwin,arshad2025simulink}. This would allow phenotyping systems to move from static trait estimation toward trajectory level understanding.

The third technology is robotics. UAVs, UGVs, robotic gantries, autonomous greenhouse systems, and robotic manipulation can turn phenotyping into an active process. Robots can collect missing evidence, revisit anomalous plants, standardize viewpoint, support under canopy imaging, and eventually execute approved interventions \cite{atefi2021robotic,chebrolu2023fieldrobot,zhang2026ground}. The fourth technology is agentic AI. Agentic systems can connect perception, retrieval, crop knowledge, digital twins, forecasting, uncertainty, and decision workflows. In the long term, this creates the possibility of self improving phenotyping platforms that continuously learn from each crop cycle, every expert correction, and each intervention outcome.

The key opportunity ahead is not to displace plant scientists, breeders, or growers, but to equip them with a fuller, more up-to-date picture of how crops are responding. Automated phenotyping will be most useful when it helps people see why a crop is falling short, what data backs that interpretation, what uncertainties still exist, and which next step should be tried. PhenoAgent is one route toward a future in which a multimodal, explainable, human-guided, and commercially viable phenotyping ecosystem evolves from simply observing to reasoning—and from reasoning to proven agricultural action.

\section{Conclusion}
\label{sec-6}
This review traced the progression of plant phenotyping from traditional trait measurements to high-throughput sensing, robotic data capture, AI-enabled analytics, and agentic crop intelligence. Classical phenotyping is still fundamental because it establishes biological trait definitions, reference values, and validation baselines; however, its dependence on manual inspection, destructive sampling, infrequent time series, and subjective ratings constrains its ability to account for dynamic crop responses shaped by seed quality, root-zone conditions, plant growth, environment, genotype, and management legacy. Contemporary automated phenotyping platforms have expanded the volume and cadence of observations using imaging, robotics, deep learning, multimodal integration, digital twins, and visualization dashboards; yet many remain measurement-centric workflows that infer traits or trigger alerts without fully interpreting crop status or informing the next decision. The key contribution of this review is the introduction of \textit{PhenoAgent} as a conceptual framework for next-generation multimodal crop intelligence. PhenoAgent is not a specific sensor, model, robot, or dashboard; instead, it is positioned as an agentic reasoning layer that links multimodal sensing, temporal fusion, crop-state representations, retrieval-augmented agronomic knowledge, uncertainty quantification, expert feedback, and recommendations for interventions across the seed-soil/substrate-plant-environment-management continuum. Under this framing, phenotyping shifts from reporting isolated trait estimates to addressing what is occurring, why it might be occurring, what evidence is absent, how confident the system is, and which management action or additional measurement should be pursued. Achieving this vision will require advances in longitudinal datasets, efficient annotation, standardized benchmarks, interoperable metadata, biological soundness, explainability, edge-cloud deployment, data governance, and translation to high-TRL systems. By coupling automated phenotyping with agentic reasoning and human-in-the-loop decision support, PhenoAgent outlines a route to interpretable, scalable, and deployment-ready crop intelligence.


\section*{Data Availability}
No new public dataset is released with this review. All datasets discussed are available through their original publications, repositories, or data providers.

\section*{Code Availability}
In this review, no validated software package is released. Conceptual PhenoAgent resources will be shared through the project repository: \href{https://github.com/Owais-CodeHub/PhenoAgent}{GitHub}.

\section*{Acknowledgements}
This research was supported by the Center for Autonomous Robotic Systems (CARS), Khalifa University of Science and Technology (KU-CARS). The authors acknowledge the use of OpenAI's ChatGPT for language and clarity improvements during manuscript preparation. The tool was used to help with grammar correction, sentence restructuring, and improving overall readability, without contributing to scientific content or analysis.

\section*{Author Contributions}
Conceptualization, Data curation, and Methodology, M.O.; Investigation and Validation, M.O. and E.I.; Writing  Original draft, M.O.; Data curation and Resources, M.O. and S.U.K., Writing review and editing, M.O., E.I. and M.U.; Supervision, Funding acquisition, and Resources, M.O., Y.A., and I.H.; All authors have read and agreed to the currently submitted version of the manuscript.

\section*{Funding}
This research was funded by the Center for Autonomous Robotic Systems (CARS), Khalifa University of Science and Technology (KU-CARS).

\section*{Competing Interest}
The authors declare no conflict of interest.


\addcontentsline{toc}{section}{References}
\bibliographystyle{naturemag}
\small
\bibliography{ref}

\end{document}